\documentclass{article}

\usepackage{microtype}
\usepackage{graphicx}
\usepackage{subfigure}
\usepackage{booktabs} % for professional tables

\usepackage{epsfig}
\usepackage{graphicx}
\usepackage{amsmath}
\usepackage{amssymb}
\usepackage{multirow}
\usepackage{enumerate}
\usepackage{bm}
\usepackage{wrapfig}
\usepackage{epstopdf}
\usepackage{wrapfig}

\DeclareMathOperator*{\argmin}{argmin}

\usepackage{cuted}
\usepackage{capt-of}

\usepackage{hyperref}

 \usepackage[accepted]{icml2023}

\usepackage{amsmath}
\usepackage{amssymb}
\usepackage{mathtools}
\usepackage{amsthm}

\usepackage[capitalize,noabbrev]{cleveref}

\theoremstyle{plain}

\theoremstyle{definition}

\theoremstyle{remark}

\usepackage[textsize=tiny]{todonotes}

\icmltitlerunning{Learning Signed Distance Functions from Noisy 3D Point Clouds via Noise to Noise Mapping}

\begin{document}

\twocolumn[
\icmltitle{Learning Signed Distance Functions from Noisy 3D Point Clouds \\
via Noise to Noise Mapping}




\begin{icmlauthorlist}
\icmlauthor{Baorui Ma}{THU}
\icmlauthor{Yu-Shen Liu}{THU}
\icmlauthor{Zhizhong Han}{WSU}
%\icmlauthor{Firstname4 Lastname4}{sch}
%\icmlauthor{Firstname5 Lastname5}{yyy}
%\icmlauthor{Firstname6 Lastname6}{sch,yyy,comp}
%\icmlauthor{Firstname7 Lastname7}{comp}
%%\icmlauthor{}{sch}
%\icmlauthor{Firstname8 Lastname8}{sch}
%\icmlauthor{Firstname8 Lastname8}{yyy,comp}
\end{icmlauthorlist}

\icmlaffiliation{THU}{School of Software, Tsinghua University, Beijing, China}
\icmlaffiliation{WSU}{Department of Computer Science, Wayne State University, Detroit, USA}
%\icmlaffiliation{sch}{School of ZZZ, Institute of WWW, Location, Country}

\icmlcorrespondingauthor{Yu-Shen Liu}{liuyushen@tsinghua.edu.cn}
%\icmlcorrespondingauthor{Firstname2 Lastname2}{first2.last2@www.uk}

\icmlkeywords{Machine Learning, ICML}

\vskip 0.3in
]



\printAffiliationsAndNotice{}  % leave blank if no need to mention equal contribution

\begin{strip}\centering
\vspace{-1.0in}
\includegraphics[width=\textwidth]{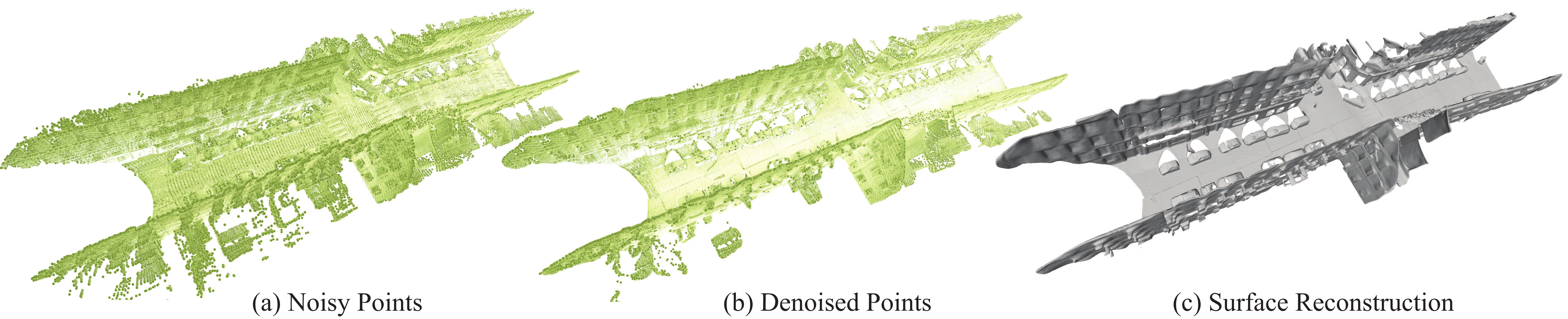}
\vspace{-0.30in}
\captionof{figure}{We introduce to learn signed distance functions (SDFs) for single noisy point clouds. Our method does not require ground truth signed distances, point normals or clean points as supervision for training. We achieve this via learning a mapping from one noisy observation to another or even on a single observation. Our novel learning manner is supported by modern Lidar systems which capture 10 to 30 noisy observations per second. We show the SDF learned from (a) a single real scan containing $10M$ points, (b) the denoised point cloud and (c) the reconstructed surface. Fig.~\ref{fig:Paris} demonstrates our superiority over the latest surface reconstructions in this case.
\vspace{-0.05in}
\label{fig:Paris1}}
\end{strip}

\begin{abstract}
   Learning signed distance functions (SDFs) from 3D point clouds is an important task in 3D computer vision. However, without ground truth signed distances, point normals or clean point clouds, current methods still struggle from learning SDFs from noisy point clouds. To overcome this challenge, we propose to learn SDFs via a noise to noise mapping, which does not require any clean point cloud or ground truth supervision for training. Our novelty lies in the noise to noise mapping which can infer a highly accurate SDF of a single object or scene from its multiple or even single noisy point cloud observations. Our novel learning manner is supported by modern Lidar systems which capture multiple noisy observations per second. We achieve this by a novel loss which enables statistical reasoning on point clouds and maintains geometric consistency although point clouds are irregular, unordered and have no point correspondence among noisy observations. Our evaluation under the widely used benchmarks demonstrates our superiority over the state-of-the-art methods in surface reconstruction, point cloud denoising and upsampling. Our code, data, and pre-trained models are available at \url{https://github.com/mabaorui/Noise2NoiseMapping/} .\vspace{-0.25in}
   %\href{https://github.com/mabaorui/Noise2NoiseMapping/}{here}.
\end{abstract}

\section{Introduction}
3D point clouds have been a popular 3D representation. We can capture 3D point clouds not only on unmanned vehicles, such as self-driving cars, but also from consumer level digital devices in our daily life, such as the iPhone. However, the raw point clouds are discretized and noisy, which is not friendly to downstream applications like virtual reality and augmented reality requiring clean surfaces. This results in a large demand of learning signed distance functions (SDFs) from 3D point clouds, since SDFs are continuous and also capable of representing arbitrary 3D topology.

%reconstructing surfaces from point clouds in various downstream applications for virtual reality and augmented reality. However, the low reconstruction accuracy is still a huge obstacle to overcome.

Deep learning based methods have shown various solutions of learning SDFs from point clouds~\cite{DBLP:conf/icml/GroppYHAL20,Atzmon_2020_CVPR,Zhizhong2021icml,jiang2020lig,Peng2021SAP}. Different from classic methods~\cite{DBLP:journals/tog/KazhdanH13,OhtakeBATS03}, they mainly leverage data-driven strategy to learn various priors from large scale dataset using deep neural networks. They usually require the signed distance ground truth~\cite{Liu2021MLS}, point normals~\cite{jiang2020lig,DBLP:conf/eccv/ChabraLISSLN20,Peng2021SAP}, additional constraints~\cite{DBLP:conf/icml/GroppYHAL20,Atzmon_2020_CVPR} or no noise assumption~\cite{Zhizhong2021icml}. These requirements significantly affect the accuracy of SDFs learned for noisy point clouds, either caused by poor generalization or the incapability of denoising. Therefore, it is still challenging to learn SDFs from noisy point clouds without clean or ground truth supervision.

To overcome this challenge, we introduce to learn SDFs from noisy point clouds via noise to noise mapping. Our method does not require ground truth signed distances and point normals or clean point clouds to learn priors. As demonstrated in Fig.~\ref{fig:Paris1}, our novelty lies in the way of learning a highly accurate SDF for a single object or scene from its several corrupted observations, i.e., noisy point clouds. Our learning manner is supported by modern Lidar systems which produce about 10 to 30 corrupted observations per second. By introducing a novel loss function containing a geometric consistency regularization, we are enabled to learn a SDF via a task of learning a mapping from one corrupted observation to another corrupted observation or even a mapping from one corrupted observation to the observation itself. The key idea of this noise to noise mapping is to leverage the statistical reasoning to reveal the uncorrupted structures upon its several corrupted observations. One of our contribution is the finding that we can still conduct statistical reasoning even there is no spatial correspondence among points on different corrupted observations. Our results achieve the state-of-the-art in different applications including surface reconstruction, point cloud denoising and upsampling under widely used benchmarks. Our contributions are listed below.\vspace{-0.15in}

\begin{enumerate}[i)]
\item We introduce a method to learn SDFs from noisy point clouds without requiring ground truth signed distances, point normals or clean point clouds.\vspace{-0.13in}
%\item We prove that we can leverage Earth Mover's Distance (EMD) to perform the statistical reasoning via noise to noise mapping and justify this idea using our novel loss function, even if 3D point clouds are irregular, unordered and have no spatial correspondence among points on different observations.
\item We prove that we can leverage Earth Mover's Distance (EMD) to perform the statistical reasoning via noise to noise mapping and justify this idea using our novel loss function, even if 3D point clouds are irregular, unordered and have no point correspondence among different observations.\vspace{-0.13in}
%\item We prove that we can leverage Earth Mover's Distance (EMD) to perform the statistical reasoning on irregular and unordered point clouds via noise to noise mapping across multiple observations without point correspondence. We also introduce a novel loss function to justify this idea using our novel loss function.
\item We achieved the state-of-the-art results in surface reconstruction, point cloud denoising and upsampling for shapes or scenes under the widely used benchmarks.\vspace{-0.1in}
\end{enumerate}

\section{Related Work}
Learning implicit functions for 3D shapes and scenes has made great progress~\cite{mildenhall2020nerf,Oechsle2021ICCV,handrwr2020,zhizhongiccv2021matching,zhizhongiccv2021completing,takikawa2021nglod,DBLP:journals/corr/abs-2105-02788,rematasICML21,feng2022np,Han2019ShapeCaptionerGCacmmm,wenxincvpr2022,li2023NeAF,ZhizhongSketch2020,wenxin_2020_CVPR,wenxin_2021a_CVPR,wenyuantip23,li2023shsnet,li2022hsurf,LP-DIF23,sayed2022simplerecon,stier2023finerecon,Shue2023triplanediffusion,DBLP:journals/corr/abs-2301-11445,gupta20233dgen,rosu2023permutosdf,zhou2022-3DOAE}. We briefly review methods with different supervision below.

\noindent\textbf{Learning from 3D Supervision. }It was explored on how to learn implicit functions, i.e., SDFs or occupancy fields, using 3D supervision including signed distances~\cite{DBLP:journals/corr/abs-1901-06802,Park_2019_CVPR,aminie2022,tianyangcvpr2022} and binary occupancy labels~\cite{MeschederNetworks,chen2018implicit_decoder}. With a condition, such as a single image~\cite{xu2019disn,pifuSHNMKL19,DBLP:conf/cvpr/ChibaneAP20,Gidi_2019_ICCV,Genova:2019:LST,seqxy2seqzeccv2020paper} or a learnable latent code~\cite{Park_2019_CVPR}, neural networks can be trained as an implicit function to model various shapes. We can also leverage point clouds as conditions~\cite{Williams_2019_CVPR,liu2020meshing,Mi_2020_CVPR,Genova:2019:LST} to learn implicit functions, and then leverage the marching cubes algorithm~\cite{Lorensen87marchingcubes} to reconstruct surfaces~\cite{jia2020learning,ErlerEtAl:Points2Surf:ECCV:2020}. To capture more detailed geometry, implicit functions are defined in local regions which are covered by voxel grids~\cite{jiang2020lig,DBLP:conf/eccv/ChabraLISSLN20,Peng2020ECCV,DBLP:journals/corr/abs-2105-02788,takikawa2021nglod,Liu2021MLS,tang2021sign}, patches~\cite{Tretschk2020PatchNets}, 3D Gaussian functions~\cite{Genova_2020_CVPR}, learnable codes~\cite{DBLP:conf/cvpr/LiWLSH22,Boulch_2022_CVPR}.

\noindent\textbf{Learning from 2D Supervision. }We can also learn implicit functions from 2D supervision, such as multiple images. The basic idea is to leverage various differentiable renderers~\cite{sitzmann2019srns,DIST2019SDFRcvpr,Jiang2019SDFDiffDRcvpr,prior2019SDFRcvpr,shichenNIPS,DBLP:journals/cgf/WuS20,Volumetric2019SDFRcvpr,lin2020sdfsrn} to render the learned implicit functions into images, so that we can obtain the error between rendered images and ground truth images. Neural volume rendering was introduced to capture the geometry and color simultaneously~\cite{mildenhall2020nerf,yariv2020multiview,yariv2021volume,geoneusfu,neuslingjie,Yu2022MonoSDF,yiqunhfSDF,Vicini2022sdf,wang2022neuris,guo2022manhattan}.

\noindent\textbf{Learning from 3D Point Clouds. }Some methods were proposed to learn implicit functions from point clouds without 3D ground truth. These methods leverage additional constraints~\cite{DBLP:conf/icml/GroppYHAL20,Atzmon_2020_CVPR,zhao2020signagnostic,atzmon2020sald,DBLP:journals/corr/abs-2106-10811,yifan2020isopoints,ben2021digs}, gradients~\cite{Zhizhong2021icml,chibane2020neural}, differentiable poisson solver~\cite{Peng2021SAP} or specially designed priors~\cite{DBLP:conf/cvpr/MaLH22,DBLP:conf/cvpr/MaLZH22} to learn signed~\cite{Zhizhong2021icml,DBLP:conf/icml/GroppYHAL20,Atzmon_2020_CVPR,zhao2020signagnostic,atzmon2020sald,chaompi2022,VisCovolume,ChaoSparse,Baoruicvpr2023} or unsigned distance fields~\cite{chibane2020neural,Zhou2022CAP-UDF}. One issue here is that they usually assume the point clouds are clean, which limits their performance in real applications due to the noise. Our method falls into this category, but we can resolve this problem using statistical reasoning via noise to noise mapping.

\noindent\textbf{Deep Learning based Point Cloud Denoising. }PointCleanNet~\cite{DBLP:journals/cgf/RakotosaonaBGMO20} was introduced to remove outliers and reduce noise from point clouds using a data-driven strategy. Graph convolution was also leveraged to reduce the noise based on dynamically constructed neighborhood graphs~\cite{DBLP:conf/eccv/PistilliFVM20}. Without supervision, TotalDenoising~\cite{DBLP:conf/iccv/Casajus0R19} inherits the same idea as Noise2Noise~\cite{DBLP:conf/icml/LehtinenMHLKAA18}. It leveraged a spatial prior term that can work for unordered point clouds. More recently, downsample-upsample architecture~\cite{DBLP:conf/mm/LuoH20} and gradient fields~\cite{luo2021score,ShapeGF} were leveraged to reduce noise. We were inspired by the idea of Noise2Noise~\cite{DBLP:conf/icml/LehtinenMHLKAA18}, our contribution lies in our finding that we can still leverage statistical reasoning among multiple noisy point clouds with specially designed losses even there is no spatial correspondence among points on different observations like the one among pixels, which is totally different from TotalDenoising~\cite{DBLP:conf/iccv/Casajus0R19}.%%\vspace{-0.15in}

\section{Method}
%%\vspace{-0.1in}
\noindent\textbf{Overview. }Given $N$ corrupted observations $S=\{\bm{N}_i|i\in[1,N],N\ge 1\}$ of an uncorrupted 3D shape or scene $\bm{S}$, we aim to learn SDFs $f$ of $\bm{S}$ from $S$ without ground truth signed distances, point normals, or clean point clouds. Here, $\bm{N}_i$ is a noisy point cloud. SDFs $f$ predicts a signed distance $d$ for an arbitrary query location $\bm{q}\in \mathbb{R}^{1\times 3}$ around $\bm{S}$, such that $d=f(\bm{q},\bm{c})$, where $\bm{c}$ is a condition denoting $\bm{S}$. We train a neural network parameterized by $\bm{\theta}$ to learn $f$, which we denote as $f_{\bm{\theta}}$. After training, we can further leverage the learned $f_{\bm{\theta}}$ for surface reconstruction, point cloud denoising, and point cloud upsampling.

\begin{figure}[tb]
  \centering
  % the following command controls the width of the embedded PS file
  % (relative to the width of the current column)
  %\includegraphics[width=.95\linewidth, bb=39 696 126 756]{figures/definition3.eps}
   \includegraphics[width=\linewidth]{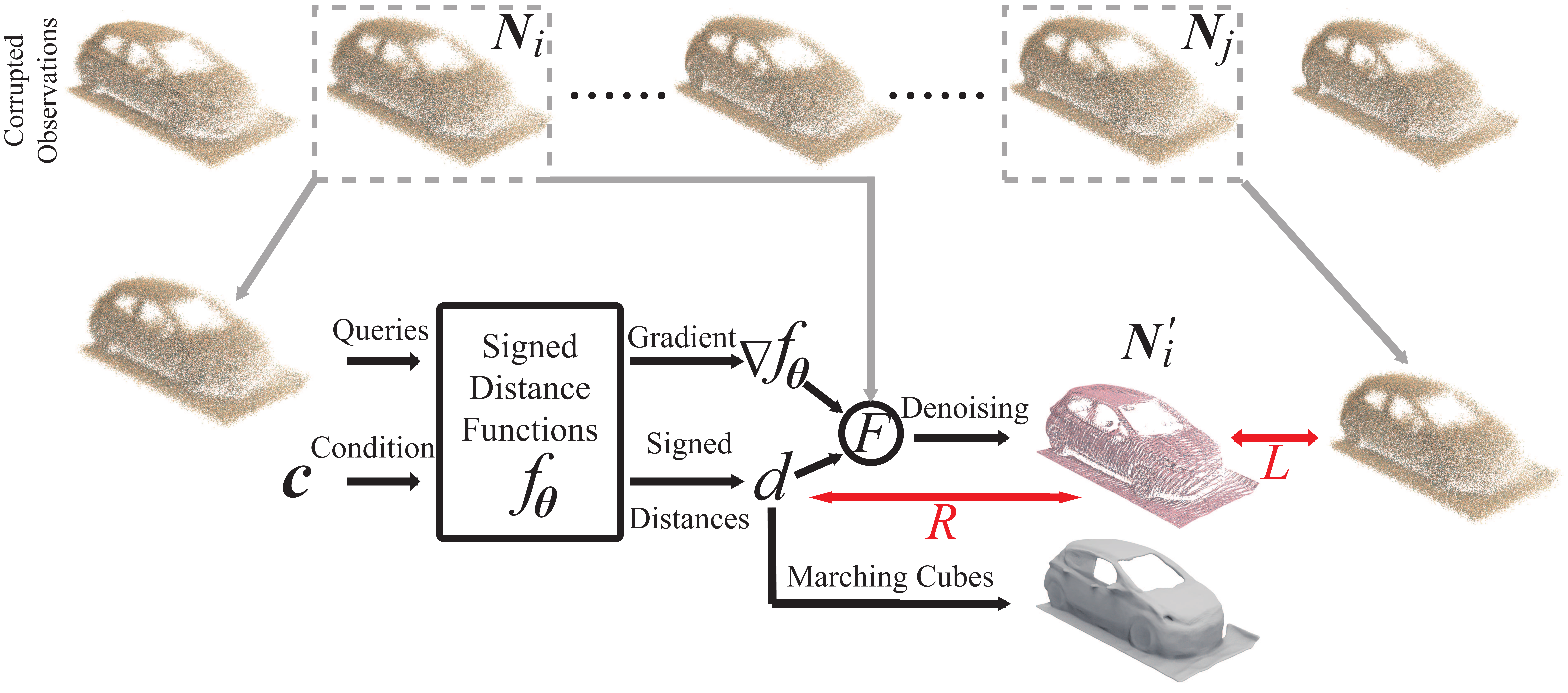}
  % replacing the above command with the one below will explicitly set
  % the bounding box of the PS figure to the rectangle (xl,yl),(xh,yh).
  % It will also prevent LaTeX from reading the PS file to determine
  % the bounding box (i.e., it will speed up the compilation process)
  % \includegraphics[width=.95\linewidth, bb=39 696 126 756]{sampleFig}
  %
  %
  \vspace{-0.38in}
\caption{\label{fig:OverviewGeneral}Given corrupted observations captured by a Lidar system per second, we learn a SDF without supervision or normals.}
\vspace{-0.25in}
\end{figure}

Our key idea of statistical reasoning is demonstrated in Fig.~\ref{fig:OverviewGeneral}. Using a noisy point cloud $\bm{N}_i$ as input, our network aims to learn SDFs $f_{\bm{\theta}}$ via learning a noise to noise mapping from $\bm{N}_i$ to another noisy point cloud $\bm{N}_j$, where $\bm{N}_j$ is also randomly selected from the corrupted observation set $S$ and $j\in[1,N]$. Our loss not only minimizes the distance between the denoised point cloud $\bm{N}_i'$ and $\bm{N}_j$ using a metric $L$ but also constrains the learned SDFs $f_{\bm{\theta}}$ to be correct using a geometric consistency regularization $R$. A denoising function $F$ conducts point cloud denoising using signed distances $d$ and gradients $\nabla f_{\bm{\theta}}$ from $f_{\bm{\theta}}$.

%Our key idea is demonstrated in Fig.~\ref{fig:OverviewGeneral}. Using a noisy point cloud $\bm{N}_i$ as input, our network aims to learn SDFs $f$ via the learning of a noise to noise mapping. The mapping generates a point cloud $\bm{N}_i'$ which is pushed to be as similar to another noisy point cloud $\bm{N}_j$ as possible, where $\bm{N}_j$ is randomly selected from the corrupted observation set $S$.

\noindent\textbf{Reducing Noise. }A common strategy for estimating the uncorrupted data from its noise corrupted observations is to find a target that has the smallest average deviation from measurements according to some loss function $L$. The data could be a scalar, a 2D image or a 3D point cloud etc.. Here, to reduce noise on point clouds, we aim to find the uncorrupted point cloud $\bm{N}'$ from its corrupted observations $\bm{N}\in S$ below,%%\vspace{-0.1in}
\begin{equation}
\label{eq:1}
\begin{aligned}
\argmin_{\bm{N}'} \mathbb{E}_{\bm{N}} \{L(\bm{N}',\bm{N})\}.
\end{aligned}
%%\vspace{-0.08in}
\end{equation}
As a conclusion of Noise2Noise~\cite{DBLP:conf/icml/LehtinenMHLKAA18} for 2D image denoising, we can learn a denoising function $F$ by pushing a denoised image $F(\bm{x})$ to be similar to as many corrupted observations $\bm{y}$ as possible, where both $\bm{x}$ and $\bm{y}$ are corrupted observations. This is an appealing conclusion since we do not need the expensive pairs of the corrupted inputs and clean targets to learn the denoising function $F$.

We want to leverage this conclusion to learn to reduce noise without requiring clean point clouds. So we transform Eq.~(\ref{eq:1}) into an equation with a denoising function $F$,%%\vspace{-0.1in}
\begin{equation}
\label{eq:2}
\begin{aligned}
\argmin_{F} \sum_{\bm{N}_i\in S}\sum_{\bm{N}_j\in S} L(F(\bm{N}_i),\bm{N}_j).
\end{aligned}
%%\vspace{-0.10in}
\end{equation}
One issue we are facing is that the conclusion of Noise2Noise may not work for 3D point clouds, due to the irregular and unordered characteristics of point clouds. For 2D images, multiple corrupted observations have the pixel correspondence. This results in an assumption that all noisy observations at the same pixel location are random realizations of a distribution around a clean pixel value. However, this assumption is invalid for point clouds. This is also the reason why TotalDenoising~\cite{DBLP:conf/iccv/Casajus0R19} does not think Eq.~(\ref{eq:1}) can work for point cloud denoising, since the noise in 3D point clouds is total. Differently, our finding is in opposite direction. We think we can still leverage Eq.~(\ref{eq:1}) to reduce noise in 3D point clouds, and the key is how to define the distance metric $L$, which is regarded as one of our contributions.

\begin{figure}[tb]
  \centering
  % the following command controls the width of the embedded PS file
  % (relative to the width of the current column)
  %\includegraphics[width=.95\linewidth, bb=39 696 126 756]{figures/definition3.eps}
   \includegraphics[width=\linewidth]{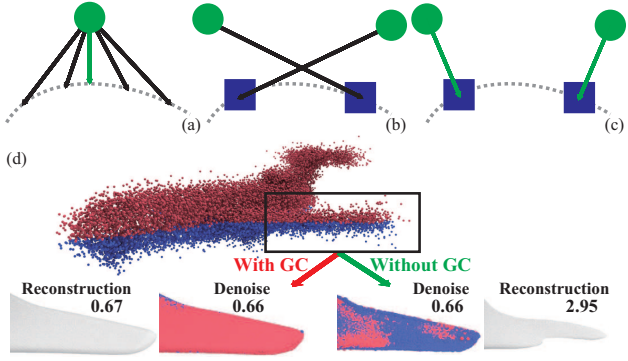}
  % replacing the above command with the one below will explicitly set
  % the bounding box of the PS figure to the rectangle (xl,yl),(xh,yh).
  % It will also prevent LaTeX from reading the PS file to determine
  % the bounding box (i.e., it will speed up the compilation process)
  % \includegraphics[width=.95\linewidth, bb=39 696 126 756]{sampleFig}
  %
  %
  \vspace{-0.2in}
\caption{\label{fig:Second}(a) Multiple paths (arrows) to pull a noise (green point) onto surface (dashed curve) but only one is the shortest (green arrows). (b) The incorrect paths (black arrows) to pull noises onto surface. (c) The expected paths (green arrows) to pull noises to points (blue square) on surface. (d) The effect of Geometric Consistency (GC).}
\vspace{-0.2in}
\end{figure}

Another issue that we are facing is how we can learn SDFs $f_{\bm{\theta}}$ via point cloud denoising in Eq.~(\ref{eq:2}). Our solution is to leverage $f_{\bm{\theta}}$ to define the denoising function $F$. This enables to conduct the learning of SDFs and point cloud denoising at the same time. Next, we will elaborate on our solutions to the aforementioned two issues.

\noindent\textbf{Denoising Function $F$. }The denoising function $F$ aims to produce a denoised point cloud $\bm{N}'$ from a noisy point cloud $\bm{N}$, so $\bm{N}'=F(\bm{N})$.

%To learn SDFs $f_{\bm{\theta}}$ of $\bm{N}$, we want the denoising procedure can also perceive the signed distance fields around $\bm{N}$. The essence of denoising is to make points floating off the surface of an object move onto the surface. There are many pathes to achieve this, but only one path is the shortest to the surface. If the moving distance is the minimum, and has a sign to indicate inside or outside of the object, this is exact signed distance. Therefore, we can leverage the signed distances predicted by SDFs $f_{\bm{\theta}}$ to denoise $\bm{N}$.

To learn SDFs $f_{\bm{\theta}}$ of $\bm{N}$, we want the denoising procedure can also perceive the signed distance fields around $\bm{N}$. The essence of denoising is to move points floating off the surface of an object onto the surface. As shown in Fig.~\ref{fig:Second} (a), there are many potential paths to achieve this, but only one path is the shortest to the surface. If we leverage this shortest path to denoise point cloud $\bm{N}$, we could involve the SDFs $f_{\bm{\theta}}$ to define the denoising function $F$, since $f_{\bm{\theta}}$ can determine the shortest path.

Here, inspired by the idea of NeuralPull~\cite{Zhizhong2021icml}, we also leverage the signed distance $d=f_{\bm{\theta}}(\bm{n},\bm{c})$ and the gradient $\nabla f_{\bm{\theta}}(\bm{n},\bm{c})$ to pull an arbitrary point $\bm{n}$ on the noisy point cloud $\bm{N}$ onto the surface. So we define the denoising function $F$ below,%\vspace{-0.07in}
\begin{equation}
\label{eq:3}
\begin{aligned}
F(\bm{n},f_{\bm{\theta}})=\bm{n}-d\times\nabla f_{\bm{\theta}}(\bm{n},\bm{c})/||\nabla f_{\bm{\theta}}(\bm{n},\bm{c})||_2.
\end{aligned}
%%\vspace{-0.05in}
\end{equation}
With Eq.~(\ref{eq:3}), we can pull all points on the noisy point cloud $\bm{N}$ onto the surface, which results in a point cloud $\bm{N}'=F(\bm{N},f_{\bm{\theta}})$. But one issue remaining is how to constrain $\bm{N}'$ to converge to the uncorrupted surface.

\noindent\textbf{Distance Metric $L$. }We investigate the distance metric $L$ so that we can constrain $\bm{N}'$ to reveal the uncorrupted surface by a statistical reasoning among the corrupted observations $S=\{\bm{N}_i\}$ using Eq.~(\ref{eq:2}). Our investigation conclusion is summarized in the following Theorem.

\noindent\textbf{Theorem 1. }\textit{Assume there was a clean point cloud $\bm{G}$ which is corrupted into observations $S=\{\bm{N}_i\}$ by sampling a noise around each point of $\bm{G}$. If we leverage EMD as the distance metric $L$ defined in Eq.~(\ref{eq:4-1}), and learn a point cloud $\bm{G}'$ by minimizing the EMD between $\bm{G}'$ and each observation in $S$, i.e., $\min_{\bm{G}'}\sum_{\bm{N}_i\in S}L(\bm{G}',\bm{N}_i)$, then $\bm{G}'$ converges to the clean point cloud $\bm{G}$, i.e., $L(\bm{G},\bm{G}')=0$.}%%%\vspace{-0.10in}
\begin{equation}
\label{eq:4-1}
\begin{aligned}
L(\bm{G},\bm{G}')=\min_{\phi:\bm{G}\to\bm{G}'}\sum_{\bm{g}\in\bm{G}}||\bm{g}-\phi(\bm{g})\|_2.
\end{aligned}
%%\vspace{-0.20in}
\end{equation}

\begin{figure}[tb]
  \centering
  % the following command controls the width of the embedded PS file
  % (relative to the width of the current column)
  %\includegraphics[width=.95\linewidth, bb=39 696 126 756]{figures/definition3.eps}
   \includegraphics[width=\linewidth]{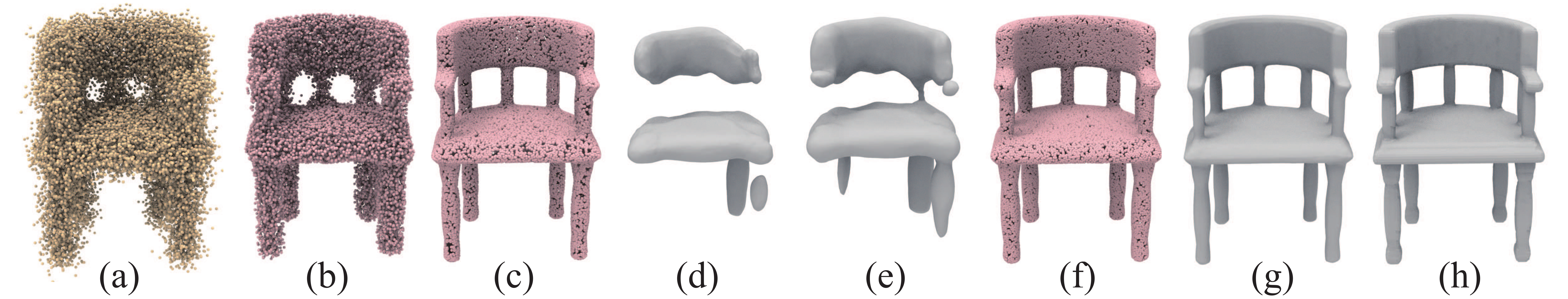}
  % replacing the above command with the one below will explicitly set
  % the bounding box of the PS figure to the rectangle (xl,yl),(xh,yh).
  % It will also prevent LaTeX from reading the PS file to determine
  % the bounding box (i.e., it will speed up the compilation process)
  % \includegraphics[width=.95\linewidth, bb=39 696 126 756]{sampleFig}
  %
  %
  \vspace{-0.3in}
\caption{\label{fig:CDEMD}The comparison with CD and EMD as the distance metric $L$ from in (b) to (e). The effect of geometric regularization in (f) and (g). (a) is noisy point cloud, (h) is the ground truth.}
\vspace{-0.25in}
\end{figure}

We prove Theorem 1 in the following appendix. We believe the one-to-one correspondence $\phi$ found in the calculation of EMD in Eq.~(\ref{eq:4-1}) plays a big role in the statistical reasoning for denoising. This is very similar to the pixel correspondence among noisy images in Noise2Noise although point clouds are irregular, unordered and have no spatial correspondence among points on different observations. We highlight this by comparing the point cloud $\bm{G}'$ optimized with EMD and Chamfer Distance (CD) as $L$ based on the same observation set $S$ in Fig.~\ref{fig:CDEMD}. Given noisy point clouds $\bm{N}_i$ like in Fig.~\ref{fig:CDEMD} (a), Fig.~\ref{fig:CDEMD} (b) demonstrates that the point cloud $\bm{G}'$ optimized with CD is still noisy, while the one optimized with EMD in Fig.~\ref{fig:CDEMD} (c) is very clean.

According to this theorem, we can learn the denoising function $F$ using Eq.~(\ref{eq:2}). $F$ produces the denoised point cloud $\bm{N}_i'=F(\bm{N}_i,f_{\bm{\theta}})$ using EMD as the distance metric $L$. This also leads to one term in our loss function below,%%\vspace{-0.1in}
\begin{equation}
\label{eq:5}
\begin{aligned}
\min_{\theta} \sum_{\bm{N}_i\in S}\sum_{\bm{N}_j\in S} L(F(\bm{N}_i,f_{\bm{\theta}}),\bm{N}_j).
\end{aligned}
%\vspace{-0.2in}
\end{equation}

\noindent\textbf{Geometric Consistency. }Although the term in Eq.~(\ref{eq:5}) can work for point cloud denoising well, as shown in Fig.~\ref{fig:CDEMD} (c), we found that the SDFs $f_{\theta}$ may not describe a correct signed distance field. With $f_{\theta}$ either learned with CD or EMD, the surfaces reconstructed using marching cubes algorithms~\cite{Lorensen87marchingcubes} in Fig.~\ref{fig:CDEMD} (d) and (e) are poor. This is because Eq.~(\ref{eq:5}) only constrains that points on the noisy point cloud should arrive onto the surface but there are no constraints on the paths to be the shortest. This is caused by the unawareness of the true surface which however is required as the ground truth by NeuralPull~\cite{Zhizhong2021icml}. The issue is further demonstrated in Fig.~\ref{fig:Second}, one situation that may happen is shown in Fig.~\ref{fig:Second} (b). With the wrong signed distances $f_{\bm{\theta}}$ and gradient $\nabla f_{\bm{\theta}}$, noises can also get pulled onto the surface, which results in a denoised point cloud with zero EMD distance to the clean point clouds. This is much different from the correct signed distance field that we expected in Fig.~\ref{fig:Second} (c).

%We further justify it by reconstructing the surface using the learned $f_{\theta}$ in Fig.~\ref{}. Fig.~\ref{} demonstrates the poor reconstructed surfaces caused by the incorrect $f_{\theta}$.

To resolve this issue, we introduce a geometric consistency to constrain $f_{\theta}$ to be correct. Our insight here is that, for an arbitrary query $\bm{n}$ around a noisy point cloud $\bm{N}_i$, the shortest distance between $\bm{n}$ and the surface can be either predicted by the SDFs $f_{\theta}$ or calculated based on the denoised point cloud $\bm{N}_i'=F(\bm{N}_i,f_{\theta})$, both of which should be consistent to each other. Therefore, the absolute value $|f_{\theta}(\bm{n},\bm{c})|$ of the signed distance predicted at $\bm{n}$ should equal to the minimum distance between $\bm{n}$ and the denoised point cloud $\bm{N}_i'=F(\bm{N}_i,f_{\theta})$. Since the point density of $\bm{N}_i'$ may slightly affect the consistency, we leverage an inequality to describe the geometric consistency,%\vspace{-0.07in}
\begin{equation}
\label{eq:6}
\begin{aligned}
|f_{\theta}(\bm{n},\bm{c})| \le \min_{\bm{n}'\in F(\bm{N}_i,f_{\theta})} ||\bm{n}-\bm{n}'||_2.
\end{aligned}
%\vspace{-0.12in}
\end{equation}
The geometric consistency is further illustrated in Fig.~\ref{fig:Second} (d). Noisy points above/below the wing can be correctly pulled onto the upper/lower surface without crossing the wing using the geometric consistency. It achieves the same denoising performance, and leads to a much more accurate SDF for surface reconstruction than the one without the geometric consistency.

\noindent\textbf{Loss Function. }With the geometric consistency, we can penalize the incorrect signed distance field shown in Fig.~\ref{fig:Second} (b) while encouraging the correct one in Fig.~\ref{fig:Second} (c). So, we leverage the geometric consistency as a regularization term $R$, which leads to our objective function below by combining Eq.~(\ref{eq:5}) and Eq.~(\ref{eq:6}),%\vspace{-0.15in}
\begin{equation}
\label{eq:7}
\begin{aligned}
%\resizebox{0.91\hsize}{!}{
\min_{\theta} \sum_{\bm{N}_i\in S}(\sum_{\bm{N}_j\in S} L(F(\bm{N}_i,f_{\theta}),\bm{N}_j)+\frac{\lambda }{|\bm{N}_i|}\sum_{\bm{n}\in\bm{N}_i}R(E),
%}
\end{aligned}
%\vspace{-0.05in}
\end{equation}
\noindent where $|\bm{N}_i|$ is the number of $\bm{n}$ on $\bm{N}_i$, $E$ is the difference defined as $(|f_{\theta}(\bm{n},\bm{c})|-\min_{\bm{n}'\in F(\bm{N}_i,f_{\theta})} ||\bm{n}-\bm{n}'||_2)$, $\lambda$ is a balance weight, and $R(E)=max(0,E)$. The effect of the geometric consistency is demonstrated in Fig.~\ref{fig:CDEMD} (f) and (g). The denoised point cloud in Fig.~\ref{fig:CDEMD} (f) shows points that are more uniformly distributed, compared with the one obtained without the geometric consistency in Fig.~\ref{fig:CDEMD} (c). More importantly, we can learn correct SDFs $f_{\theta}$ to reconstruct plausible surface in Fig.~\ref{fig:CDEMD} (g), compared to the one obtained without the geometric consistency in Fig.~\ref{fig:CDEMD} (e) and the ground truth in Fig.~\ref{fig:CDEMD} (h).

\begin{table*}[t]
\centering
%\caption{Denoising comparison. L2CD$\times 10^4$ and P2M $\times 10^4$.}  % ????????
\resizebox{\linewidth}{!}{
%\makebox[\linewidth]{
    \begin{tabular}{c|c|cccccc|cccccc}  % ?????
     \toprule
        \multicolumn{2}{c|}{Point Number}&\multicolumn{6}{c|}{10K(Sparse)}&\multicolumn{6}{c}{50K(Dense)}  \\
        \hline
        \multicolumn{2}{c|}{Noise}&\multicolumn{2}{c}{1\%}&\multicolumn{2}{c}{2\%}&\multicolumn{2}{c|}{3\%}&\multicolumn{2}{c}{1\%}&\multicolumn{2}{c}{2\%}&\multicolumn{2}{c}{3\%}  \\
        &Model&CD&P2M&CD&P2M&CD&P2M&CD&P2M&CD&P2M&CD&P2M\\
    \hline
        \multirow{9}{*}{\rotatebox{90}{PU}}&\multicolumn{1}{l|}{Bilateral}&3.646&1.342&5.007&2.018&6.998&3.557&0.877&0.234&2.376&1.389&6.304&4.730 \\
        &\multicolumn{1}{l|}{Jet}&2.712&0.613&4.155&1.347&6.262&2.921&0.851&0.207&2.432&1.403&5.788&4.267\\
        &\multicolumn{1}{l|}{MRPCA}&2.972&0.922&3.728&1.117&5.009&1.963&0.669&\textbf{0.099}&2.008&1.003&5.775&4.081\\
        &\multicolumn{1}{l|}{GLR}&2.959&1.052&3.773&1.306&4.909&2.114&0.696&0.161&1.587&0.830&3.839&2.707\\
        \cline{2-14}
         &\multicolumn{1}{l|}{PCNet}&3.515&1.148&7.469&3.965&13.067&8.737&1.049&0.346&1.447&0.608&2.289&1.285\\
         &\multicolumn{1}{l|}{GPDNet}&3.780&1.337&8.007&4.426&13.482&9.114&1.913&1.037&5.021&3.736&9.705&7.998\\
         &\multicolumn{1}{l|}{DMR}&4.482&1.722&4.982&2.115&5.892&2.846&1.162&0.469&1.566&0.800&2.632&1.528\\
         &\multicolumn{1}{l|}{SBP}&2.521&0.463&3.686&1.074&4.708&1.942&0.716&0.150&1.288&0.566&1.928&1.041\\
         \cline{2-14}
         &\multicolumn{1}{l|}{TTD-Un}&3.390&0.826&7.251&3.485&13.385&8.740&1.024&0.314&2.722&1.567&7.474&5.729\\
          &\multicolumn{1}{l|}{SBP-Un}&3.107&0.888&4.675&1.829&7.225&3.726&0.918&0.265&2.439&1.411&5.303&3.841\\
         \cline{2-14}
         &\multicolumn{1}{l|}{\textbf{Ours}}&\textbf{1.060}&\textbf{0.241}&\textbf{2.925}&\textbf{1.010}&\textbf{4.221}&\textbf{1.847}
         &\textbf{0.377}&0.155&\textbf{1.029}&\textbf{0.484}&\textbf{1.654}&\textbf{0.972}\\
    \toprule
    \toprule
    \multirow{9}{*}{\rotatebox{90}{PC}}&\multicolumn{1}{l|}{Bilaterall}&4.320&1.351&6.171&1.646&8.295&2.392&1.172&0.198&2.478&0.634&6.077&2.189 \\
        &\multicolumn{1}{l|}{Jet}&3.032&0.830&5.298&1.372&7.650&2.227&1.091&0.180&2.582&0.700&5.787&2.144\\
        &\multicolumn{1}{l|}{MRPCA}&3.323&0.931&4.874&1.178&6.502&1.676&0.966&0.140&2.153&0.478&5.570&1.976\\
        &\multicolumn{1}{l|}{GLR}&3.399&0.956&5.274&1.146&7.249&1.674&0.964&0.134&2.015&0.417&4.488&1.306\\
        \cline{2-14}
         &\multicolumn{1}{l|}{PCNet}&3.849&1.221&8.752&3.043&14.525&5.873&1.293&0.289&1.913&0.505&3.249&1.076\\
         &\multicolumn{1}{l|}{GPDNet}&5.470&1.973&10.006&3.650&15.521&6.353&5.310&1.716&7.709&2.859&11.941&5.130\\
         &\multicolumn{1}{l|}{DMR}&6.602&2.152&7.145&2.237&8.087&2.487&1.566&0.350&2.009&0.485&2.993&0.859\\
         &\multicolumn{1}{l|}{SBP}&3.369&0.830&5.132&1.195&6.776&1.941&1.066&0.177&1.659&0.354&2.494&\textbf{0.657}\\
         \cline{2-14}
         &\multicolumn{1}{l|}{\textbf{Ours}}&\textbf{2.047}&\textbf{0.518}&\textbf{2.056}&\textbf{0.519}&\textbf{5.331}&\textbf{1.935}
         &\textbf{0.426}&\textbf{0.129}&\textbf{1.043}&\textbf{0.316}&\textbf{2.22}&1.096\\
       \toprule
   \end{tabular}%}
   }\vspace{-0.18in}
   \caption{Denoising comparison. L2CD$\times 10^4$ and P2M $\times 10^4$.}
   \label{table:denoising}
   \vspace{-0.15in}
\end{table*}

\noindent\textbf{More Details. }We sample more queries around the input noisy point cloud $\bm{N}_i$ using the method introduced in NeuralPull~\cite{Zhizhong2021icml}. We randomly sample a batch of $B$ queries as input, and also randomly sample the same number of points from another noisy point cloud $\bm{N}_j$ as target. Using batches enables us to process large scale point clouds,  makes it possible to leverage noisy point clouds with different point numbers even we use EMD as the distance metric $L$, and more importantly, does not affect the performance. We train $f_{\bm{\theta}}$ to overfit to a single shape or scene or overfit to multiple shapes or scenes using conditions $\bm{c}$ to indicate different shapes or scenes.

\begin{figure*}[tb]
  \centering
  % the following command controls the width of the embedded PS file
  % (relative to the width of the current column)
  %\includegraphics[width=.95\linewidth, bb=39 696 126 756]{figures/definition3.eps}
   \includegraphics[width=\linewidth]{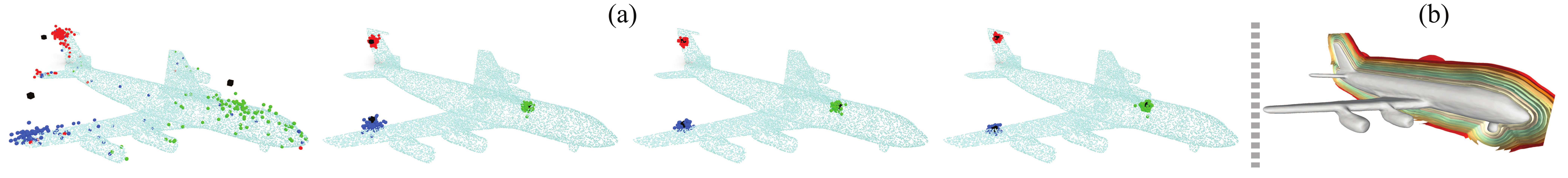}
  % replacing the above command with the one below will explicitly set
  % the bounding box of the PS figure to the rectangle (xl,yl),(xh,yh).
  % It will also prevent LaTeX from reading the PS file to determine
  % the bounding box (i.e., it will speed up the compilation process)
  % \includegraphics[width=.95\linewidth, bb=39 696 126 756]{sampleFig}
  %
  %
  \vspace{-0.3in}
\caption{\label{fig:NoiseOpt}(a) Visualization of optimization in $4$ epochs via noise to noise mapping. $3$ queries (black cubes) sampled from one noisy point cloud get pulled onto the surface. For each query, we minimize its distance to all targets (in the same color) matched from another noisy point cloud by the mapping $\phi$ in metric $L$. More details can be found in our video. (b) Surface reconstruction and multiple level-sets.}
\vspace{-0.18in}
\end{figure*}

We visualize the optimization process in $4$ epochs in Fig.~\ref{fig:NoiseOpt} (a). We show how the $3$ queries (black cubes) get pulled progressively onto the surface (Cyan). For each query, we also show its corresponding target in each one of $100$ batches in the same color (red, green, blue), and each target is established by the mapping $\phi$ in the metric $L$. The essence of statical reasoning in each epoch is that each query will be pulled to the average point of all targets from all batches since the distance between the query and each target should be minimized. Although the targets are found all over the shape in the first epoch, the targets surround the query more tightly as the query gets pulled to the surface in the following epochs. This makes queries get pulled onto the surface which results in an accurate SDF visualized in the surface reconstruction and level-sets in Fig.~\ref{fig:NoiseOpt} (b).

\noindent\textbf{One Noisy Point Cloud. }Although we prove Theorem $1$ based on multiple noisy point clouds ($N>1$), we surprisingly found that our method can also work well when only one noisy point cloud ($N=1$) is available. Specifically, we regard the queries sampled around the noisy point cloud $\bm{N}_i$ as input and regard $\bm{N}_i$ as target. We believe the reason why $N=1$ works is that the knowledge learned via statistical reasoning in the batch based training can be well generalized to various regions. We will report our results learned from multiple or one noisy point clouds in experiments.

%\noindent\textbf{Noise Types. }We do not assume any noise types. More details can be found in our proof of Theorem 1. So, we work well with different noise types in Fig.~\ref{fig:assumption}. In evaluations, we follow the same noise types in benchmarks or deal with unknown ones in real scans for fair comparisons.%\vspace{-0.15in}

\begin{figure}[htb]
 %\vspace{-0.2in}
  \centering
  % the following command controls the width of the embedded PS file
  % (relative to the width of the current column)
  %\includegraphics[width=.95\linewidth, bb=39 696 126 756]{figures/definition3.eps}
   \includegraphics[width=\linewidth]{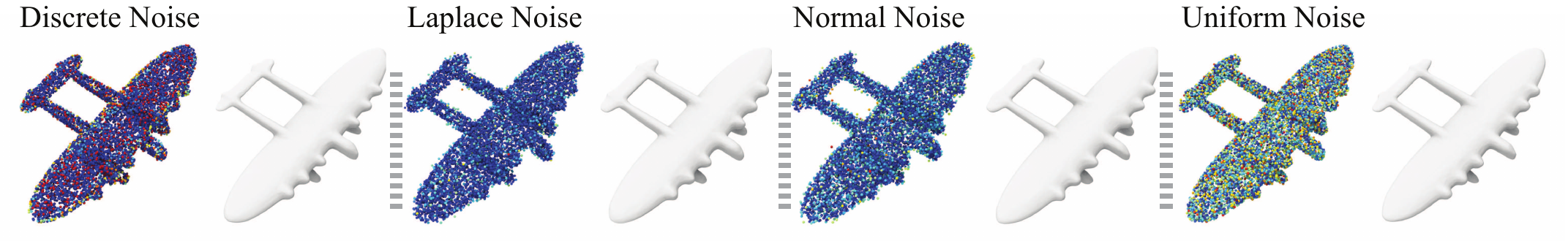}
  % replacing the above command with the one below will explicitly set
  % the bounding box of the PS figure to the rectangle (xl,yl),(xh,yh).
  % It will also prevent LaTeX from reading the PS file to determine
  % the bounding box (i.e., it will speed up the compilation process)
  % \includegraphics[width=.95\linewidth, bb=39 696 126 756]{sampleFig}
  %
  %
  \vspace{-0.3in}
\caption{\label{fig:assumption}Reconstruction with different kinds of noises.}
\vspace{-0.22in}
\end{figure}

\noindent\textbf{Noise Types. }We work well with different types of noises in Fig.~\ref{fig:assumption}. We use zero-mean noises in our proof of Theorem 1, but we find we work well with unknown noises in real scans in experiments. In evaluations, we also use the same type of noises in benchmarks for fair comparisons.%\vspace{-0.15in}

\section{Experiments and Analysis}
%\vspace{-0.1in}
We evaluate our method in two steps. We first evaluate our method in applications that only care about points, such as point cloud denoising and upsampling. So, we only leverage Eq.~(\ref{eq:5}) to produce the denoised or upsampled point clouds. Then, we evaluate our method trained with the loss in Eq.~(\ref{eq:7}) in surface reconstruction, where $\lambda=0.1$.%\vspace{-0.15in}

\begin{table}[t]
\centering
%\caption{Upsampling comparison. L1CD$\times 10^4$ and P2M $\times 10^4$.}  % ????????
\resizebox{\linewidth}{!}{
%\makebox[\linewidth]{
    \begin{tabular}{c|ccc|ccc}  % ?????
     \toprule
     \multirow{1}{*}{Points}&\multicolumn{3}{c|}{5K}&\multicolumn{3}{c}{10K} \\
     \hline
     &PU-Net&SBP&\textbf{Ours}&PU-Net&SBP&\textbf{Ours}\\
     \hline
     CD&3.445&1.696&\textbf{0.592}&2.862&1.454&\textbf{0.418}\\
     P2M&1.669&0.295&\textbf{0.156}&1.166&0.181&\textbf{0.155}\\
     \toprule
\end{tabular}%}
   }\vspace{-0.2in}
   \caption{Upsampling comparison. L2CD$\times 10^4$ and P2M $\times 10^4$.}
   \label{table:upsampling}
   \vspace{-0.2in}
\end{table}

\subsection{Point Cloud Denoising}
\begin{figure}[tb]
  \centering
  % the following command controls the width of the embedded PS file
  % (relative to the width of the current column)
  %\includegraphics[width=.95\linewidth, bb=39 696 126 756]{figures/definition3.eps}
   \includegraphics[width=\linewidth]{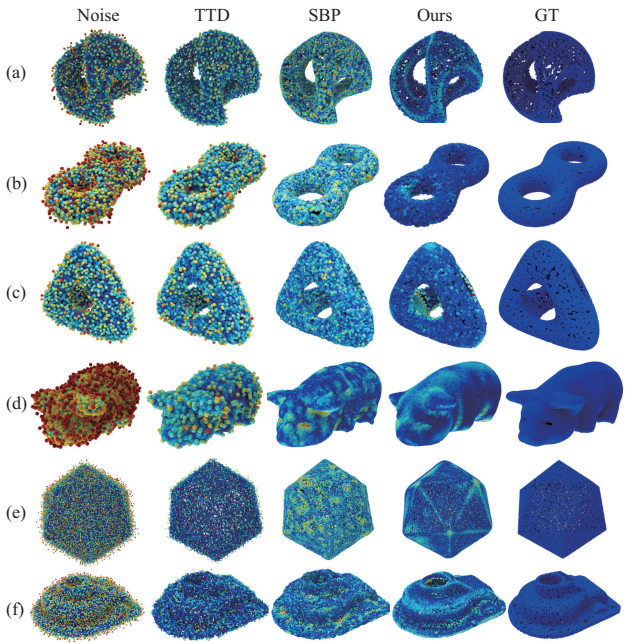}
  % replacing the above command with the one below will explicitly set
  % the bounding box of the PS figure to the rectangle (xl,yl),(xh,yh).
  % It will also prevent LaTeX from reading the PS file to determine
  % the bounding box (i.e., it will speed up the compilation process)
  % \includegraphics[width=.95\linewidth, bb=39 696 126 756]{sampleFig}
  %
  %
  \vspace{-0.3in}
\caption{\label{fig:denoise}Visual comparison in point cloud denoising. Error at each point is shown in color. (a) and (b) $10K$ points with $3\%$ noise. (c) $10K$ points with $2\%$ noise. (d) and (e) $50K$ points with $3\%$ noise. (f) $50K$ points with with $2\%$ noise.}
\vspace{-0.15in}
\end{figure}

%\vspace{-0.1in}
\noindent\textbf{Dataset and Metric. }For the fair comparison with the state-of-the-art results, we follow SBP~\cite{luo2021score} to evaluate our method under two benchmarks named as PU and PC that were released by PUNet~\cite{DBLP:conf/cvpr/YuLFCH18} and PointCleanNet~\cite{DBLP:journals/cgf/RakotosaonaBGMO20}. We report our results under 20 shapes in the test set of PU and 10 shapes in the test set of PC. We use Poisson disk to sample $10K$ and $50K$ points from each shape respectively as the ground truth clean point clouds in two different resolutions. The clean point cloud is normalized into the unit sphere. In each resolution, we add Gaussian noise with three standard deviations including $1\%$, $2\%$, $3\%$ to the clean point clouds. We leverage L2 Chamfer Distance (L2CD) and point to mesh distance (P2M) to evaluate the denoising performance.
For each test shape, we generate $N=200$ noisy point clouds to train our method. We sample $B=250$ points in each batch. We report our results and numerical comparison in Tab.~\ref{table:denoising}. The compared methods include Bilateral~\cite{DBLP:journals/tog/FleishmanDC03}, Jet~\cite{DBLP:journals/cagd/CazalsP05}, MRPCA~\cite{DBLP:journals/cgf/MatteiC17}, GLR~\cite{DBLP:journals/tip/ZengCNPY20}, PCNet~\cite{DBLP:journals/cgf/RakotosaonaBGMO20}, GPDNet~\cite{DBLP:conf/eccv/PistilliFVM20}, DMR~\cite{DBLP:conf/mm/LuoH20}, TTD~\cite{DBLP:conf/iccv/Casajus0R19}, and SBP~\cite{luo2021score}. These methods require learned priors and can not directly use multiple observations. The comparison with different conditions indicates that our method significantly outperforms traditional point cloud denoising methods and deep learning based point cloud denoising methods in both supervised and unsupervised (``-Un'') settings. Error map comparison with TTD~\cite{DBLP:conf/iccv/Casajus0R19} and SBP~\cite{luo2021score} in Fig.~\ref{fig:denoise} further demonstrates our state-of-the-art denoising performance.%\vspace{-0.15in}

\subsection{Point Cloud Upsampling}
%\vspace{-0.1in}
\noindent\textbf{Dataset and Metric. }We use the PU dataset mentioned before to evaluate the $f_{\bm{\theta}}$ learned in our denoising experiments in point cloud upsampling. Following SBP~\cite{luo2021score}, we produce an upsampled point cloud with an upsampling rate of 4 from a sparse point cloud by denoising the sparse point cloud with noise. We compare the denoised point cloud and the ground truth, and report L2CD and P2M comparison in Tab.~\ref{table:upsampling}. We compared with PU-Net~\cite{DBLP:conf/cvpr/YuLFCH18} and SBP~\cite{luo2021score}. The comparison demonstrates that our method can perform the statistical reasoning to reveal points on the surface more accurately.%\vspace{-0.15in}

\subsection{Surface Reconstruction for Shapes}
%\vspace{-0.1in}
\noindent\textbf{ShapeNet. }We first report our surface reconstruction performance under the test set of 13 classes in ShapeNet~\cite{shapenet2015}. The train and test splits follow COcc~\cite{DBLP:conf/eccv/PengNMP020}. Following IMLS~\cite{Liu2021MLS}, we leverage point clouds with $3000$ points as clean truth, and add Gaussian noise with a standard deviation of 0.005. For each clean point cloud, we generate $N=200$ noisy point clouds with a batch size of $B=3000$. We leverage L1 Chamfer Distance (L1CD), Normal Consistency (NC)~\cite{MeschederNetworks}, and F-score~\cite{Tatarchenko_2019_CVPR} with a threshold of $1\%$ as metrics.

\begin{figure}[tb]
  \centering
  % the following command controls the width of the embedded PS file
  % (relative to the width of the current column)
  %\includegraphics[width=.95\linewidth, bb=39 696 126 756]{figures/definition3.eps}
   \includegraphics[width=\linewidth]{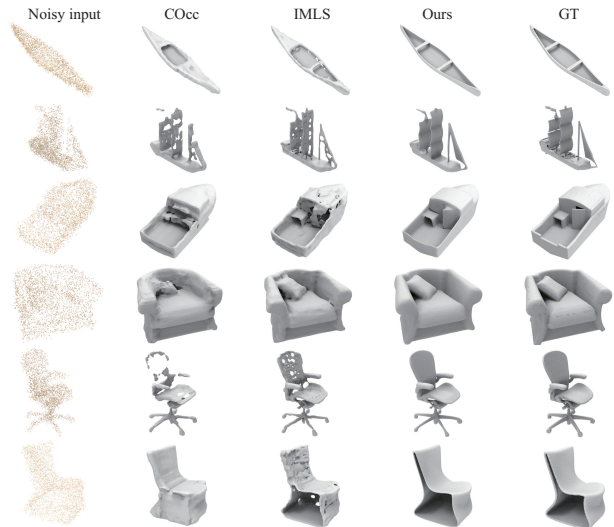}
  % replacing the above command with the one below will explicitly set
  % the bounding box of the PS figure to the rectangle (xl,yl),(xh,yh).
  % It will also prevent LaTeX from reading the PS file to determine
  % the bounding box (i.e., it will speed up the compilation process)
  % \includegraphics[width=.95\linewidth, bb=39 696 126 756]{sampleFig}
  %
  %
  \vspace{-0.35in}
\caption{\label{fig:ShapeNet}Comparison in surface reconstruction under ShapeNet.}
\vspace{-0.15in}
\end{figure}

We compare our methods with methods including PSR~\cite{DBLP:journals/tog/KazhdanH13}, PSG~\cite{DBLP:conf/cvpr/FanSG17}, R2N2~\cite{DBLP:conf/eccv/ChoyXGCS16}, Atlas~\cite{groueix2018papier}, COcc~\cite{DBLP:conf/eccv/PengNMP020}, SAP~\cite{Peng2021SAP}, OCNN~\cite{wang2020deep}, IMLS~\cite{Liu2021MLS} and POCO~\cite{Boulch_2022_CVPR}. The numerical comparison in Tab.~\ref{table:shapenet3} demonstrates our state-of-the-art surface reconstruction accuracy over 13 classes. Although we do not require the ground truth supervision, our method outperforms the supervised methods such as SAP~\cite{Peng2021SAP}, COcc~\cite{DBLP:conf/eccv/PengNMP020} and IMLS~\cite{Liu2021MLS}. We further demonstrate our superiority in the reconstruction of complex geometry in the visual comparison in Fig.~\ref{fig:ShapeNet}. More numerical and visual comparisons can be found in the following appendix.

\begin{table}[t]
\centering
%\caption{L1CD$\times 10$ comparison under ShapeNet.}  % ????????
\resizebox{\linewidth}{!}{
    \begin{tabular}{c|cccccccccc}  % ?????
     \toprule
     %\multirow{2}{*}{category}&\multicolumn{9}{c}{Chamfer-\textit{$L_1$} $\times 10$} \\
     &PSR&PSG&R2N2&Atlas&COcc&SAP&OCNN&IMLS&POCO&\textbf{Ours}\\
     \hline
     L1CD$\times 10$&0.299&0.147&0.173&0.093&0.044&0.034&0.067&0.031&0.030&\textbf{0.026}\\
     NC&0.772&-&0.715&0.855&0.938&0.944&0.932&0.944&0.950&\textbf{0.962}\\
     F-Score&0.612&0.259&0.400&0.708&0.942&0.975&0.800&0.983&0.984&\textbf{0.991}\\
     \toprule
\end{tabular}}
\vspace{-0.1in}
\caption{L1CD, NC and F-Score comparison under ShapeNet.}
\vspace{-0.25in}
   \label{table:shapenet3}
\end{table}

\noindent\textbf{FAMOUS and ABC. }We further evaluate our method using the test set in FAMOUS and ABC dataset provided by P2S~\cite{ErlerEtAl:Points2Surf:ECCV:2020}. The clean point cloud is corrupted with noise at different levels. We follow NeuralPull~\cite{Zhizhong2021icml} to report L2 Chamfer Distance (L2CD). Different from previous experiments, we only leverage single $N=1$ noisy point clouds to train our method with a batch size of $B=1000$.

\begin{figure}[tb]
\vspace{-0.20in}
  \centering
  % the following command controls the width of the embedded PS file
  % (relative to the width of the current column)
  %\includegraphics[width=.95\linewidth, bb=39 696 126 756]{figures/definition3.eps}
   \includegraphics[width=\linewidth]{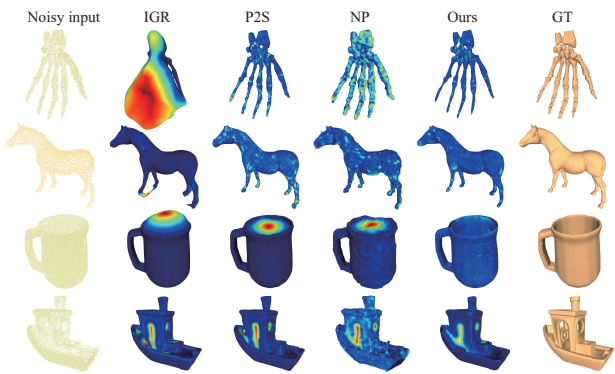}
  % replacing the above command with the one below will explicitly set
  % the bounding box of the PS figure to the rectangle (xl,yl),(xh,yh).
  % It will also prevent LaTeX from reading the PS file to determine
  % the bounding box (i.e., it will speed up the compilation process)
  % \includegraphics[width=.95\linewidth, bb=39 696 126 756]{sampleFig}
  %
  %
  \vspace{-0.3in}
\caption{\label{fig:famous}Visual comparison in surface reconstruction under FAMOUS. Point to surface error at each vertex is shown in color.}
\vspace{-0.25in}
\end{figure}

We compare our methods with methods including DSDF~\cite{Park_2019_CVPR}, Atlas~\cite{groueix2018papier}, PSR~\cite{DBLP:journals/tog/KazhdanH13}, P2S~\cite{ErlerEtAl:Points2Surf:ECCV:2020}, NP~\cite{Zhizhong2021icml}, IMLS~\cite{Liu2021MLS}, PCP~\cite{DBLP:conf/cvpr/MaLZH22}, POCO~\cite{Boulch_2022_CVPR}, and OnSF~\cite{DBLP:conf/cvpr/MaLH22}. The comparison in Tab.~\ref{table:NOX3noise} demonstrates that our method can reveal more accurate surfaces from noisy point clouds even we do not have training set, ground truth supervision or even multiple noisy point clouds. The statistical reasoning on point clouds and geometric regularization produce more accurate surfaces as demonstrated by the error map comparison under FAMOUS in Fig.~\ref{fig:famous}.

\begin{table}[tb]
\centering
%\caption{L2CD$\times100$ comparison under ABC and Famous.}  %
\resizebox{\linewidth}{!}{
    \begin{tabular}{c|c|c|c|c|c|c|c|c|c|c}  % 32=0.1S隆锚?64=0.2S隆锚?128=1.3S * 600000
     \hline
          Dataset&DSDF&Atlas&PSR&P2S&NP& IMLS&PCP&POCO&OnSF&Ours\\   %
     \hline
       ABC var& 12.51 & 4.04 & 3.29& 2.14&0.72 & 0.57&0.49 &2.01&3.52&\textbf{0.113} \\ %0.4806
       ABC max& 11.34 & 4.47 & 3.89& 2.76&1.24& 0.68& 0.57&2.50&4.30&\textbf{0.139} \\ %0.4806
       \hline
       F-med &9.89&4.54&1.80&1.51&0.28& 0.80&0.07&1.50&0.59&\textbf{0.033}\\%sparse:0.593
       F-max &13.17&4.14&3.41&2.52&0.31& 0.39 &0.30&2.75&3.64&\textbf{0.117}\\%sparse:3.638
     \hline
   \end{tabular}}
   \vspace{-0.15in}
   \caption{L2CD$\times100$ comparison under ABC and Famous.}
   \label{table:NOX3noise}
   \vspace{-0.25in}
\end{table}

\noindent\textbf{D-FAUST and SRB. }Finally, we evaluate our method under the real scanning dataset D-FAUST~\cite{dfaust:CVPR:2017} and SRB~\cite{Williams_2019_CVPR}. We follow SAP~\cite{Peng2021SAP} to evaluate our result using L1CD, NC~\cite{MeschederNetworks}, and F-score~\cite{Tatarchenko_2019_CVPR} with a threshold of $1\%$ using the same set of shapes. We use single $N=1$ noisy point clouds to train our method with a batch size of $B=5000$.

\begin{figure}[tb]
  \centering
  % the following command controls the width of the embedded PS file
  % (relative to the width of the current column)
  %\includegraphics[width=.95\linewidth, bb=39 696 126 756]{figures/definition3.eps}
   \includegraphics[width=\linewidth]{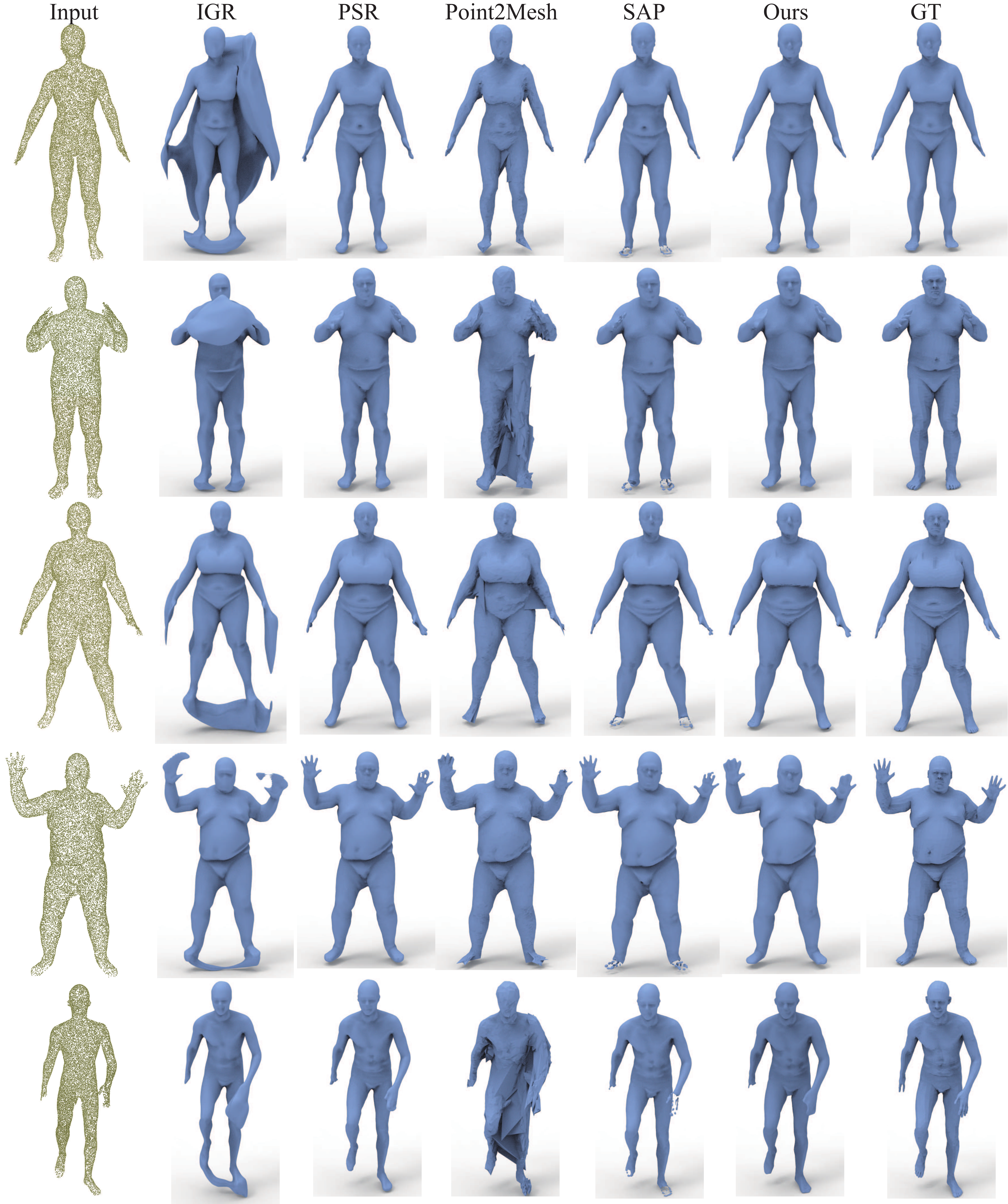}
  % replacing the above command with the one below will explicitly set
  % the bounding box of the PS figure to the rectangle (xl,yl),(xh,yh).
  % It will also prevent LaTeX from reading the PS file to determine
  % the bounding box (i.e., it will speed up the compilation process)
  % \includegraphics[width=.95\linewidth, bb=39 696 126 756]{sampleFig}
  %
  %
  \vspace{-0.3in}
\caption{\label{fig:DFAUST}Comparison in surface reconstruction under DFAUST.}
\vspace{-0.3in}
\end{figure}

\begin{figure*}[htb]
%\vspace{-0.15in}
  \centering
  % the following command controls the width of the embedded PS file
  % (relative to the width of the current column)
  %\includegraphics[width=.95\linewidth, bb=39 696 126 756]{figures/definition3.eps}
   \includegraphics[width=0.8\linewidth]{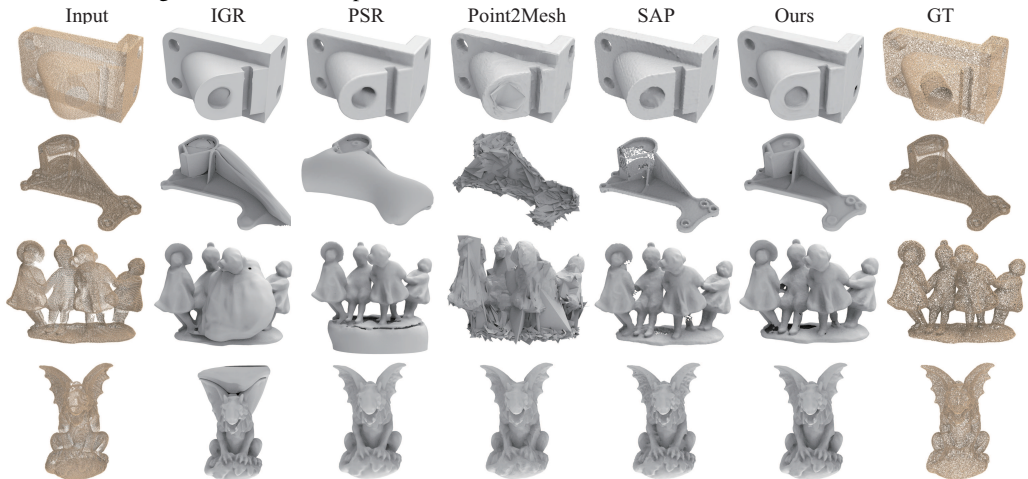}
  % replacing the above command with the one below will explicitly set
  % the bounding box of the PS figure to the rectangle (xl,yl),(xh,yh).
  % It will also prevent LaTeX from reading the PS file to determine
  % the bounding box (i.e., it will speed up the compilation process)
  % \includegraphics[width=.95\linewidth, bb=39 696 126 756]{sampleFig}
  %
  %
  \vspace{-0.2in}
\caption{\label{fig:SRB}Comparison in surface reconstruction under SRB.}
\vspace{-0.2in}
\end{figure*}

We compare our methods with the methods including IGR~\cite{DBLP:conf/icml/GroppYHAL20}, Point2Mesh~\cite{DBLP:journals/tog/HanockaMGC20}, PSR~\cite{DBLP:journals/tog/KazhdanH13}, SAP~\cite{Peng2021SAP}. We report numerical comparison in Tab.~\ref{table:dfaust} and Tab.~\ref{table:srb}. Although we only do statistical reasoning on a single noisy point cloud and do not require point normals as SAP~\cite{Peng2021SAP}, our method still handles the noise in real scanning well, which achieves much smoother and more accurate structure. The comparison in Fig.~\ref{fig:DFAUST} and Fig.~\ref{fig:SRB} shows that our method can produce more accurate surfaces without missing parts on both rigid and non-rigid shapes.
%\begin{table}[t]
%\centering
%\caption{Comparison in surface reconstruction under D-FAUST.}  % ????????
%\resizebox{\linewidth}{!}{
%%\makebox[\linewidth]{
%    \begin{tabular}{c|ccc}  % ?????
%     \toprule
%     Method&L1CD$\times 10$&F-Score&Normal Consistency\\
%     \hline
%     IGR~\cite{DBLP:conf/icml/GroppYHAL20}&0.235&0.805&0.911\\
%     Point2Mesh~\cite{DBLP:journals/tog/HanockaMGC20}&0.071&0.855&0.905\\
%     SPSR~\cite{DBLP:journals/tog/KazhdanH13}&0.044&0.966&0.965\\
%     SAP~\cite{Peng2021SAP}&0.043&0.966&0.959\\
%     \textbf{Ours}&\textbf{0.037}&\textbf{0.996}&\textbf{0.970}\\
%     \toprule
%\end{tabular}%}
%   }
%   \label{table:dfaust}
%\end{table}

\begin{table}[t]
\centering
%\caption{Comparison in surface reconstruction under D-FAUST.}  % ????????
\resizebox{0.8\linewidth}{!}{
%\makebox[\linewidth]{
    \begin{tabular}{c|ccccc}  % ?????
     \toprule
     %Method&L1CD$\times 10$&F-Score&Normal Consistency\\
     Metrics&IGR&Point2Mesh&PSR&SAP&\textbf{Ours}\\
     \hline
     L1CD$\times 10$&0.235&0.071&0.044&0.043&\textbf{0.037}\\
     F-Score&0.805&0.855&0.966&0.966&\textbf{0.996}\\
     NC&0.911&0.905&0.965&0.959&\textbf{0.970}\\
%     IGR~\cite{DBLP:conf/icml/GroppYHAL20}&0.235&0.805&0.911\\
%     Point2Mesh~\cite{DBLP:journals/tog/HanockaMGC20}&0.071&0.855&0.905\\
%     SPSR~\cite{DBLP:journals/tog/KazhdanH13}&0.044&0.966&0.965\\
%     SAP~\cite{Peng2021SAP}&0.043&0.966&0.959\\
%     \textbf{Ours}&\textbf{0.037}&\textbf{0.996}&\textbf{0.970}\\
     \toprule
\end{tabular}%}
   }
   \vspace{-0.1in}
   \caption{Comparison in surface reconstruction under D-FAUST.}
   \vspace{-0.15in}
   \label{table:dfaust}
\end{table}

%\begin{table}[t]
%\centering
%\caption{Comparison in surface reconstruction under SRB.}  % ????????
%\resizebox{\linewidth}{!}{
%%\makebox[\linewidth]{
%    \begin{tabular}{c|ccc}  % ?????
%     \toprule
%     Method&L1CD$\times 10$&F-Score&Normal Consistency\\
%     \hline
%     IGR~\cite{DBLP:conf/icml/GroppYHAL20}&0.178&0.755\\
%     Point2Mesh~\cite{DBLP:journals/tog/HanockaMGC20}&0.116&0.648\\
%     SPSR~\cite{DBLP:journals/tog/KazhdanH13}&0.232&0.735\\
%     SAP~\cite{Peng2021SAP}&0.076&0.830\\
%     \textbf{Ours}&\textbf{0.067}&\textbf{0.835}\\
%     \toprule
%\end{tabular}%}
%   }
%   \label{table:srb}
%\end{table}

\begin{table}[t]
\centering
%\caption{Comparison in surface reconstruction under SRB.}  % ????????
\resizebox{0.8\linewidth}{!}{
%\makebox[\linewidth]{
    \begin{tabular}{c|ccccc}  % ?????
     \toprule
     %Method&L1CD$\times 10$&F-Score&Normal Consistency\\
     Metrics&IGR&Point2Mesh&PSR&SAP&\textbf{Ours}\\
     \hline
     L1CD$\times 10$&0.178&0.116&0.232&0.076&\textbf{0.067}\\
     F-Score&0.755&0.648&0.735&0.830&\textbf{0.835}\\
%     IGR~\cite{DBLP:conf/icml/GroppYHAL20}&0.178&0.755\\
%     Point2Mesh~\cite{DBLP:journals/tog/HanockaMGC20}&0.116&0.648\\
%     SPSR~\cite{DBLP:journals/tog/KazhdanH13}&0.232&0.735\\
%     SAP~\cite{Peng2021SAP}&0.076&0.830\\
%     \textbf{Ours}&\textbf{0.067}&\textbf{0.835}\\
     \toprule
\end{tabular}%}
   }
   \vspace{-0.1in}
   \caption{Comparison in surface reconstruction under SRB.}
   \vspace{-0.1in}
   \label{table:srb}
\end{table}

\subsection{Surface Reconstruction for Scenes}
%\vspace{-0.1in}
%\noindent\textbf{3D Scene. }We evaluate our method under real scene scanning dataset~\cite{DBLP:journals/tog/ZhouK13}. We sample $1000$ points per $m^2$ from Lounge and Copyroom, and only leverage $N=1$ noisy point cloud to train our method with a batch size of $B=5000$. We leverage the pretrained models of COcc~\cite{DBLP:conf/eccv/PengNMP020} and LIG~\cite{jiang2020lig} and retrain NP~\cite{Zhizhong2021icml} and DeepLS~\cite{DBLP:conf/eccv/ChabraLISSLN20} to produce their results with the same input. We also provide LIG and DeepLS with the ground truth point normals. Numerical comparison in Tab.~\ref{table:t12} demonstrates that our method significantly outperforms the state-of-the-art. Fig.~\ref{fig:Scenes} further demonstrates that we can produce much smoother surfaces with more geometry details.

\begin{figure*}[tb]
  \centering
  % the following command controls the width of the embedded PS file
  % (relative to the width of the current column)
  %\includegraphics[width=.95\linewidth, bb=39 696 126 756]{figures/definition3.eps}
   \includegraphics[width=\linewidth]{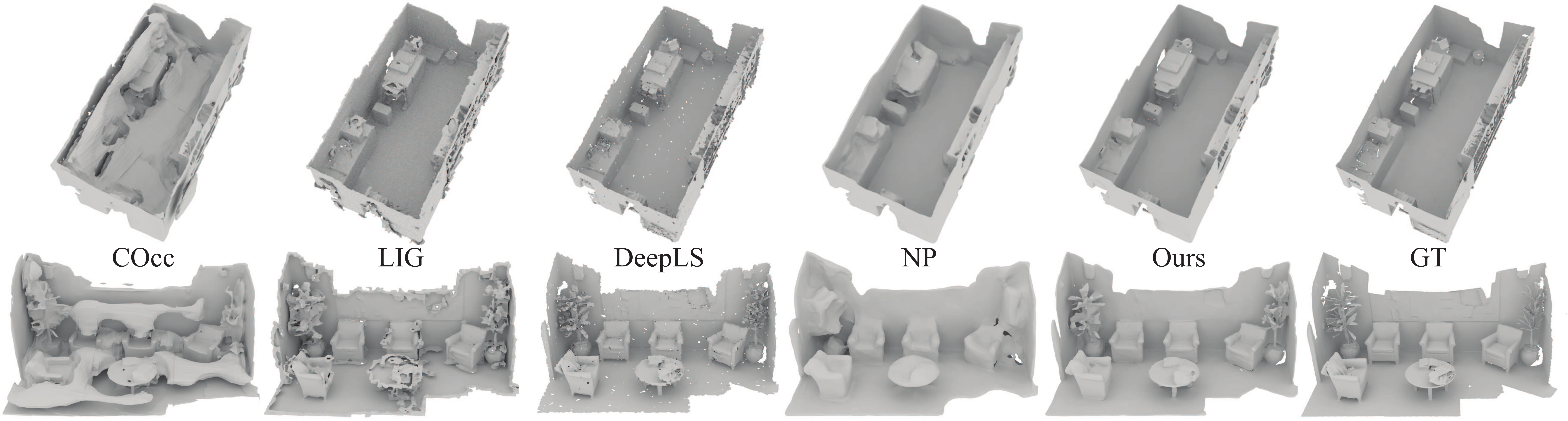}
  % replacing the above command with the one below will explicitly set
  % the bounding box of the PS figure to the rectangle (xl,yl),(xh,yh).
  % It will also prevent LaTeX from reading the PS file to determine
  % the bounding box (i.e., it will speed up the compilation process)
  % \includegraphics[width=.95\linewidth, bb=39 696 126 756]{sampleFig}
  %
  %
  \vspace{-0.34in}
\caption{\label{fig:Scenes}Visual comparison in surface reconstruction under 3D Scene dataset.}
\vspace{-0.15in}
\end{figure*}

\noindent\textbf{3D Scene. }We evaluate our method under real scene scan dataset~\cite{DBLP:journals/tog/ZhouK13}. We sample $1000$ points per $m^2$ from Lounge and Copyroom, and only leverage $N=1$ noisy point cloud to train our method with a batch size of $B=5000$. We leverage the pretrained models of COcc and LIG and retrain NP and DeepLS to produce their results with the same input. We also provide LIG and DeepLS with the ground truth point normals. Numerical comparison in Tab.~\ref{table:t12} demonstrates that our method significantly outperforms the state-of-the-art. Fig.~\ref{fig:Scenes} further demonstrates that we can produce much smoother surfaces with more geometry details.

\begin{table}[t]
%%\vspace{-0.38in}
\centering
%\caption{Surface reconstruction under 3D Scene.L2-CD$\times 1000$. Norm is short for normal. The unit of error is mm.}  % ????????Comparison of shape reconstruction with known camera pose from silhouette images with different resolutions in terms of CD
\resizebox{\linewidth}{!}{
    \begin{tabular}{c|c|c|c||c|c|c}
     \hline
       %\cline{1-12}
       %\hline
        %Metric& MPU& DeepLS& LIG & ConvOCC & Ours&MPU& DeepLS& LIG & ConvOCC & Ours&MPU& DeepLS& LIG & ConvOCC & Ours \\  % ?????隆矛篓垄?

        &\multicolumn{3}{c||}{Lounge}&\multicolumn{3}{c}{Copyroom}\\
        \hline
       &L2CD&L1CD&NC&L2CD&L1CD&NC\\
     \hline
     COcc~\cite{DBLP:conf/eccv/PengNMP020}&9.540 &0.046 &0.894 &10.97 &0.045 &0.892\\
     LIG~\cite{jiang2020lig}&9.672 &0.056 &0.833 &3.61 &0.036 &0.810\\
     DeepLS~\cite{DBLP:conf/eccv/ChabraLISSLN20}&6.103&0.053&0.848&0.609&0.021&0.901\\
     NP~\cite{Zhizhong2021icml}&1.079 &0.019 &0.910 &5.795 &0.036 &0.862\\
     \hline
     Ours&\textbf{0.602}&\textbf{0.016}&\textbf{0.923}&\textbf{0.442}&\textbf{0.016}&\textbf{0.903}\\
     \hline
   \end{tabular}}
   \vspace{-0.15in}
   \caption{Surface reconstruction under 3D Scene dataset. L2-CD$\times 10^3$. The unit of error is mm.}
   \vspace{-0.25in}
   \label{table:t12}
\end{table}

\vspace{-0.1in}

\begin{figure}[t]
  \centering
  % the following command controls the width of the embedded PS file
  % (relative to the width of the current column)
  %\includegraphics[width=.95\linewidth, bb=39 696 126 756]{figures/definition3.eps}
   \includegraphics[width=\linewidth]{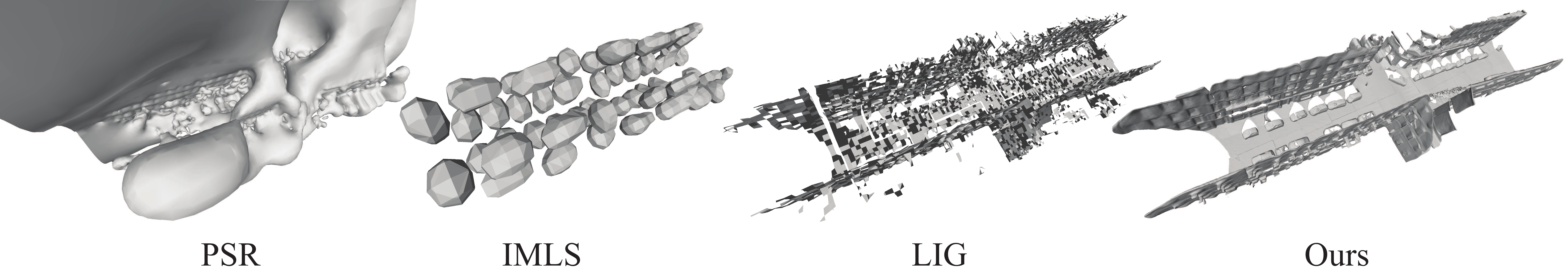}
  % replacing the above command with the one below will explicitly set
  % the bounding box of the PS figure to the rectangle (xl,yl),(xh,yh).
  % It will also prevent LaTeX from reading the PS file to determine
  % the bounding box (i.e., it will speed up the compilation process)
  % \includegraphics[width=.95\linewidth, bb=39 696 126 756]{sampleFig}
  %
  %
  \vspace{-0.35in}
\caption{\label{fig:Paris}Comparison in surface reconstruction from real scans.}
\vspace{-0.2in}
\end{figure}

\begin{figure}[htb]
  \centering
  % the following command controls the width of the embedded PS file
  % (relative to the width of the current column)
  %\includegraphics[width=.95\linewidth, bb=39 696 126 756]{figures/definition3.eps}
   \includegraphics[width=\linewidth]{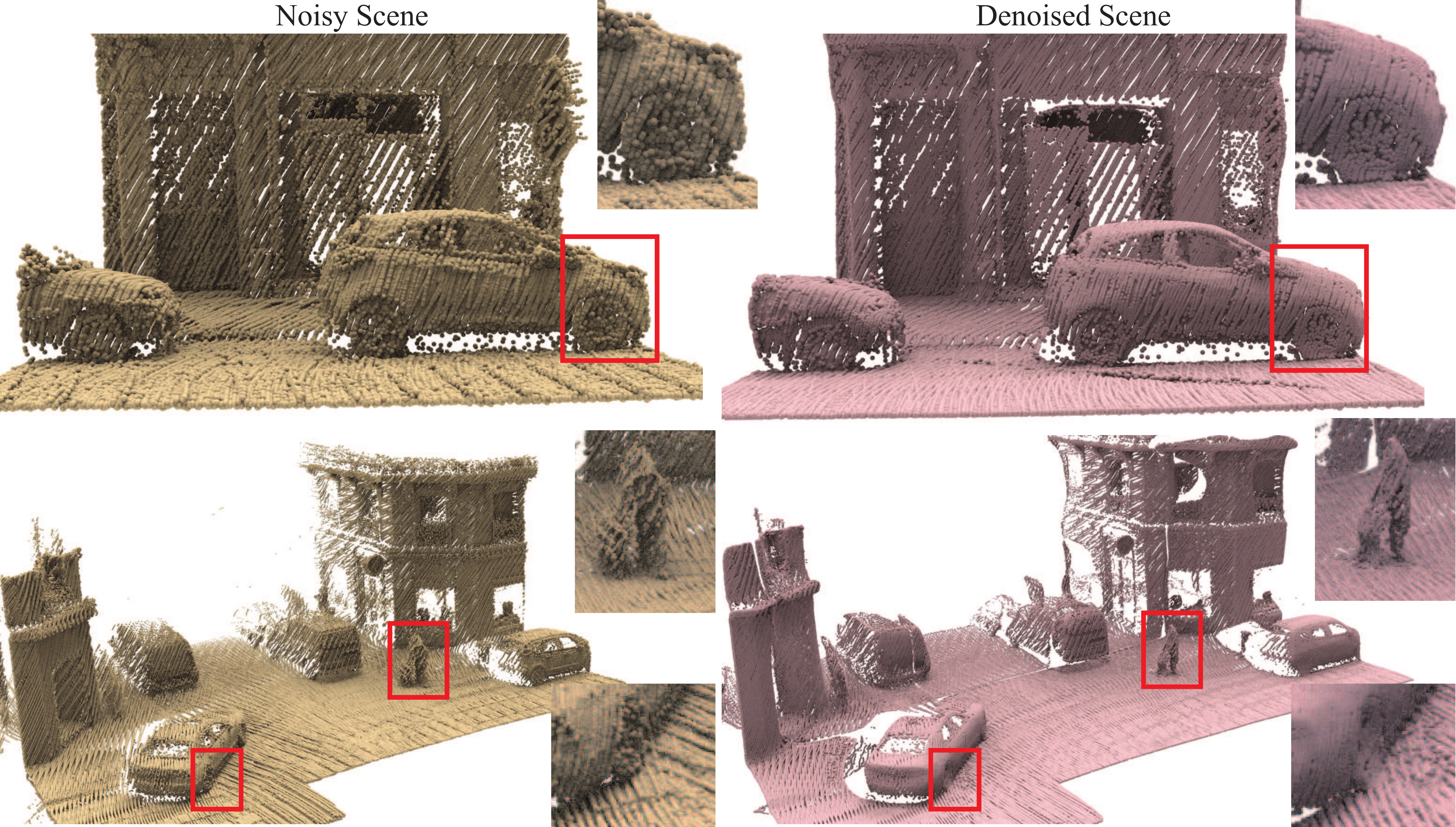}
  % replacing the above command with the one below will explicitly set
  % the bounding box of the PS figure to the rectangle (xl,yl),(xh,yh).
  % It will also prevent LaTeX from reading the PS file to determine
  % the bounding box (i.e., it will speed up the compilation process)
  % \includegraphics[width=.95\linewidth, bb=39 696 126 756]{sampleFig}
  %
  %
  \vspace{-0.33in}
\caption{\label{fig:ParisZoom}Demonstration of denoising on real scans.}
\vspace{-0.18in}
\end{figure}

\noindent\textbf{Paris-rue-Madame. }We further evaluate our method under another real scene scan dataset~\cite{DBLP:conf/icpram/SernaMGD14}. We only use $N=1$ noisy point cloud with a batch size of $B=5000$. We split the $10M$ points into $50$ chunks each of which is used to learn a SDF. Similarly, we use each chunk to evaluate IMLS~\cite{Liu2021MLS} and LIG~\cite{jiang2020lig} with their pretrained models. Our superior performance over the latest methods in large scale surface reconstruction is demonstrated in Fig.~\ref{fig:Paris}. Our denoised point clouds in a smaller scene are detailed in Fig.~\ref{fig:ParisZoom}.%\vspace{-0.15in}

\begin{table}[htb]
\centering
\vspace{-0.19in}
%\caption{Effect of batch size $B$ under PU.}  % ????????
\resizebox{\linewidth}{!}{
    \begin{tabular}{c|c|c|c|c|c|c}  % 32=0.1S隆锚?64=0.2S隆锚?128=1.3S * 600000
     \hline
          $B$ & 100 & 250 & 1000 & 2000 & 5000 & 10000\\   % ?????隆矛篓垄?
     \hline
       L2CD$\times10^4$&12.398 &\textbf{4.221}&4.578&5.628&5.998&6.217\\ %0.4806
       P2M$\times10^4$&5.482 &\textbf{1.847}&1.901&2.112&2.221&2.342\\
       \hline
   \end{tabular}}
   \vspace{-0.15in}
   \caption{Effect of batch size $B$ under PU.}
   \vspace{-0.1in}
   \label{table:NOX14}
\end{table}

\subsection{Ablation Studies}
%\vspace{-0.1in}
We conduct ablation studies under the test set of PU. We first explore the effect of batch size $B$, training iterations, and the number $N$ of noisy point clouds in point cloud denoising. Tab.~\ref{table:NOX14} indicates that more points in each batch will slow down the convergence. Tab.~\ref{table:NOX15} demonstrates that more training iterations help perform statistical reasoning better to remove noise. Tab.~\ref{table:NOX16} indicates that more corrupted observations are the key to increase the performance of statistical reasoning although one corrupted observation is also fine to perform statistical reasoning well.

We further highlight the effect of EMD as the distance metric $L$ and geometric consistency regularization $R$ in denoising and surface reconstruction in Tab.~\ref{table:NOX17}. The comparison shows that we can not perform statistical reasoning on point clouds using CD, and EMD can only reveal the surface in statistical reasoning for denoising but not learn meaningful signed distance fields without $R$. Moreover, we found the $\lambda$ weighting $R$ slightly affects our performance. More additional studies are in the following appendix. %\vspace{-0.1in}

\begin{table}[!]
\centering
%\caption{Number of training iterations under PU.}  % ????????
%\resizebox{\linewidth}{!}{
    \begin{tabular}{c|c|c|c|c}  % 32=0.1S隆锚?64=0.2S隆锚?128=1.3S * 600000
     \hline
          %Iterations $\times 10^4$& 100 & 80 & 60 & 40 \\   % ?????隆矛篓垄?
          Iterations $\times 10^4$& 40 & 60 & 80 & 100 \\   % ?????隆矛篓垄?
     \hline
%       L2CD$\times10^4$&4.224 &\textbf{4.221}&4.364&4.887\\ %0.4806
%       P2M$\times10^4$&1.849 &\textbf{1.847}&1.885&2.032\\
       L2CD$\times10^4$&4.887&4.364&\textbf{4.221}&4.224\\ %0.4806
       P2M$\times10^4$&2.032&1.885&\textbf{1.847}&1.849\\
       \hline
   \end{tabular}%}
   \vspace{-0.1in}
   \caption{Number of training iterations under PU.}
   \vspace{-0.2in}
   \label{table:NOX15}
\end{table}

\begin{table}[!]
\centering
%\caption{Effect of $N$ under PU.}  % ????????
\resizebox{\linewidth}{!}{
    \begin{tabular}{c|c|c|c|c|c|c|c}  % 32=0.1S隆锚?64=0.2S隆锚?128=1.3S * 600000
     \hline
          $N$& 1& 2 & 10 & 20 & 50 &100 &200\\   % ?????隆矛篓垄?
     \hline
       L2CD$\times10^4$&4.976&4.898&4.665&4.558&4.432&4.224 &\textbf{4.221}\\ %0.4806
       P2M$\times10^4$&2.132&2.079&1.997&1.996&1.899&\textbf{1.847} &\textbf{1.847}\\
       \hline
   \end{tabular}}
   \vspace{-0.15in}
   \caption{Effect of $N$ under PU.}
   \vspace{-0.1in}
   \label{table:NOX16}
\end{table}

\begin{table}[!]
\centering
%\caption{Effect of CD and EMD as the distance metric $L$ and geometry consistency regularization $R$ under PU. L2CD$\times10^4$.}  % ????????
\resizebox{\linewidth}{!}{
    \begin{tabular}{c|c|c|c|c|c}  % 32=0.1S隆锚?64=0.2S隆锚?128=1.3S * 600000
     \hline
          &CD&EMD,$\lambda=0$&EMD,$\lambda=0.05$&EMD,$\lambda=0.1$&EMD,$\lambda=0.2$\\% ?????隆矛篓垄?
     \hline
       Denoise&73.786&\textbf{4.221}&4.245&4.252&4.832\\ %0.4806
       Reconstruction&81.573&80.917&5.721&\textbf{4.277}&4.993\\
       \hline
   \end{tabular}}
   \vspace{-0.2in}
   \caption{Effect of CD and EMD as the distance metric $L$ and geometry consistency regularization $R$ under PU. L2CD$\times10^4$.}
   \vspace{-0.3in}
   \label{table:NOX17}
\end{table}

\section{Conclusion}
\vspace{-0.1in}
\label{Conclusion}
We introduce to learn SDFs from noisy point clouds via noise to noise mapping. We explore the feasibility of learning SDFs from multiple noisy point clouds or even one noisy point cloud without the ground truth signed distances, point normals or clean point clouds. Our noise to noise mapping enables the statistical reasoning on point clouds although there is no spatial correspondence among points on different noisy point clouds. Our key insight in noise to noise mapping is to use EMD as the metric in the statistical reasoning. With the capability of the statistical reasoning, we successfully reveal surfaces from noisy point clouds by learning highly accurate SDFs. We evaluate our method under synthetic dataset or real scanning dataset for both shapes or scenes. The effectiveness of our method is justified by our state-of-the-art performance in different applications.%\vspace{-0.15in}

\vspace{-0.15in}
\section*{Acknowledgement}
\vspace{-0.1in}
We thank reviewers who gave useful comments. This work was supported by National Key R\&D Program of China (2022YFC3800600), the National Natural Science Foundation of China (62272263, 62072268), and in part by Tsinghua-Kuaishou Institute of Future Media Data.

%One limitation is that we did not explore how the point density of noisy point clouds affects the performance. This would be a good direction for future work.

\newpage
\bibliography{papers}
\bibliographystyle{icml2023}

%%%%%%%%%%%%%%%%%%%%%%%%%%%%%%%%%%%%%%%%%%%%%%%%%%%%%%%%%%%%%%%%%%%%%%%%%%%%%%%
%%%%%%%%%%%%%%%%%%%%%%%%%%%%%%%%%%%%%%%%%%%%%%%%%%%%%%%%%%%%%%%%%%%%%%%%%%%%%%%
% APPENDIX
%%%%%%%%%%%%%%%%%%%%%%%%%%%%%%%%%%%%%%%%%%%%%%%%%%%%%%%%%%%%%%%%%%%%%%%%%%%%%%%
%%%%%%%%%%%%%%%%%%%%%%%%%%%%%%%%%%%%%%%%%%%%%%%%%%%%%%%%%%%%%%%%%%%%%%%%%%%%%%%
\newpage
\appendix
\twocolumn

\section{Network Architectures}
We employ a network that is modified based on OccNet~\cite{MeschederNetworks}. Since the output of OccNet is a value with a range of [0,1], we replace the sigmoid function that produces this output with the tanh function, which can output a signed distance value with a range of [-1,1], where the sign indicates the inside or outside of the 3D shape. In addition, we also replace the Resblock used in OccNet by simple fully connected layers to simplify the OccNet, which highlights the advantage of our method.

\section{Query Sampling}
We sample more queries around a noisy point cloud if there is only one noisy point cloud available. We leverage a method introduced by NeuralPull~\cite{Zhizhong2021icml} to sample queries around each point on the noisy point cloud.

\section{Surface Reconstruction}
\noindent\textbf{Numerical Comparison. }We report more detailed comparison under ShapeNet~\cite{shapenet2015}. Due to the text limit in the main body, we only report the mean metric over all 13 classes under ShapeNet. We compare our methods with methods including PSR~\cite{DBLP:journals/tog/KazhdanH13}, PSG~\cite{DBLP:conf/cvpr/FanSG17}, R2N2~\cite{DBLP:conf/eccv/ChoyXGCS16}, Atlas~\cite{groueix2018papier}, COcc~\cite{DBLP:conf/eccv/PengNMP020}, SAP~\cite{Peng2021SAP}, OCNN~\cite{wang2020deep}, and IMLS~\cite{Liu2021MLS}. We report the numerical comparison in terms of L1CD, NC, and F-score in Tab.~\ref{table:shapenet3app}, Tab.~\ref{table:shapenet1app}, and Tab.~\ref{table:shapenet2app}, respectively.

\begin{table*}[htb]
\centering
%\caption{L1CD$\times 10$ comparison under ShapeNet.}  % ????????
\resizebox{\linewidth}{!}{
    \begin{tabular}{c|cccccccccc}  % ?????
     \toprule
     %\multirow{2}{*}{category}&\multicolumn{9}{c}{Chamfer-\textit{$L_1$} $\times 10$} \\
     &PSR&PSG&R2N2&Atlas&COcc&SAP&OCNN&IMLS&POCO&\textbf{Ours}\\
     \hline
     airplane&0.437&0.102&0.151&0.064&0.034&0.027&0.063&0.025&0.023&\textbf{0.022}\\
     bench&0.544&0.128&0.153&0.073&0.035&0.032&0.065&0.030&0.028&\textbf{0.025}\\
     cabinet&0.154&0.164&0.167&0.112&0.047&0.037&0.071&0.035&0.037&\textbf{0.034}\\
     car&0.180&0.132&0.197&0.099&0.075&0.045&0.077&0.040&0.041&\textbf{0.037}\\
     chair&0.369&0.168&0.181&0.114&0.046&0.036&0.066&0.035&0.033&\textbf{0.026}\\
     display&0.280&0.160&0.170&0.089&0.036&0.030&0.066&0.029&0.028&\textbf{0.022}\\
     lamp&0.278&0.207&0.243&0.137&0.059&0.047&0.067&0.031&0.033&\textbf{0.027}\\
     speaker&0.148&0.205&0.199&0.142&0.063&0.041&0.073&0.040&0.041&\textbf{0.033}\\
     rifle&0.409&0.091&0.167&0.051&0.028&0.023&0.062&0.021&0.019&\textbf{0.019}\\
     sofa&0.227&0.144&0.160&0.091&0.041&0.032&0.066&0.031&0.030&\textbf{0.027}\\
     table&0.393&0.166&0.177&0.102&0.038&0.033&0.066&0.032&0.031&\textbf{0.028}\\
     telephone&0.281&0.110&0.130&0.054&0.027&0.023&0.061&0.023&0.022&\textbf{0.017}\\
     vessele&0.181&0.130&0.169&0.078&0.043&0.030&0.064&0.027&0.025&\textbf{0.024}\\
     \hline
     mean&0.299&0.147&0.173&0.093&0.044&0.034&0.067&0.031&0.030&\textbf{0.026}\\
     \toprule
\end{tabular}}%\vspace{-0.15in}
\caption{L1CD$\times 10$ comparison under ShapeNet.}
   \label{table:shapenet3app}
\end{table*}%%\vspace{-0.2in}

\begin{table*}[htb]
\centering
%\caption{NC comparison under ShapeNet.}  % ????????
\resizebox{\linewidth}{!}{
%\makebox[\linewidth]{
    \begin{tabular}{c|cccccccccc}  % ?????
     \toprule
     %\multirow{2}{*}{category}&\multicolumn{9}{c}{Normal Consistency} \\
     %&PSR&PSGN&R2N2&AtlasNet&ConvONet&SAP&OCNNC&IMLS&\textbf{Ours(256)}\\
     &PSR&PSG&R2N2&Atlas&COcc&SAP&OCNN&IMLS&POCO&\textbf{Ours}\\
     \hline
     airplane&0.747&-&0.669&0.854&0.931&0.931&0.918&0.937&0.944&\textbf{0.960}\\
     bench&0.649&-&0.691&0.820&0.921&0.920&0.914&0.922&0.928&\textbf{0.935}\\
     cabinet&0.835&-&0.786&0.875&0.956&0.957&0.941&0.955&0.961&\textbf{0.975}\\
     car&0.783&-&0.719&0.827&0.893&0.897&0.867&0.882&0.894&\textbf{0.937}\\
     chair&0.715&-&0.673&0.829&0.943&0.952&0.941&0.950&0.956&\textbf{0.965}\\
     display&0.749&-&0.747&0.905&0.968&0.972&0.960&0.973&0.975&\textbf{0.981}\\
     lamp&0.765&-&0.598&0.759&0.900&0.921&0.911&0.922&0.929&\textbf{0.957}\\
     speaker&0.843&-&0.735&0.867&0.938&0.950&0.936&0.947&0.952&\textbf{0.977}\\
     rifle&0.788&-&0.700&0.837&0.929&0.937&0.932&0.943&\textbf{0.949}&0.938\\
     sofa&0.826&-&0.754&0.888&0.958&0.963&0.949&0.963&0.967&\textbf{0.978}\\
     table&0.706&-&0.734&0.867&0.959&0.962&0.946&0.962&0.966&\textbf{0.970}\\
     telephone&0.805&-&0.847&0.957&0.983&0.984&0.974&0.984&0.985&\textbf{0.987}\\
     vessele&0.820&-&0.641&0.837&0.918&0.930&0.922&0.932&0.940&\textbf{0.951}\\
     \hline
     mean&0.772&-&0.715&0.855&0.938&0.944&0.932&0.944&0.950&\textbf{0.962}\\
     \toprule
\end{tabular}%}
   }%\vspace{-0.15in}
   \caption{NC comparison under ShapeNet.}
   \label{table:shapenet1app}%\vspace{-0.2in}
\end{table*}

\begin{figure}[htb]
  \centering
  % the following command controls the width of the embedded PS file
  % (relative to the width of the current column)
  %\includegraphics[width=.95\linewidth, bb=39 696 126 756]{figures/definition3.eps}
   \includegraphics[width=\linewidth]{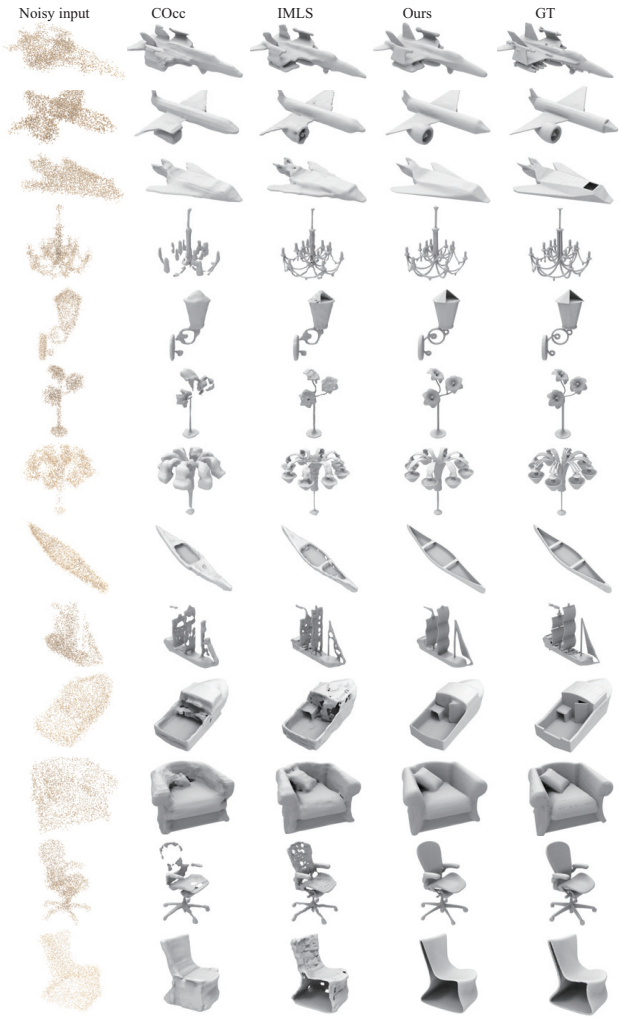}
  % replacing the above command with the one below will explicitly set
  % the bounding box of the PS figure to the rectangle (xl,yl),(xh,yh).
  % It will also prevent LaTeX from reading the PS file to determine
  % the bounding box (i.e., it will speed up the compilation process)
  % \includegraphics[width=.95\linewidth, bb=39 696 126 756]{sampleFig}
  %
  %
  %%\vspace{-0.35in}
\caption{\label{fig:ShapeNet1}Visual comparison with COcc~\cite{DBLP:conf/eccv/PengNMP020} and IMLS~\cite{Liu2021MLS} in surface reconstruction under ShapeNet.}
%%\vspace{-0.28in}
\end{figure}

\noindent\textbf{Visual Comparison. }We report more surface reconstruction results under ShapeNet~\cite{shapenet2015} in Fig.~\ref{fig:ShapeNet11}, Fig.~\ref{fig:ShapeNet1} and  Fig.~\ref{fig:ShapeNet2}. This comparison demonstrates that our method can reconstruct more geometry details than the state-of-the-art methods.

\begin{figure}[tb]
  \centering
  % the following command controls the width of the embedded PS file
  % (relative to the width of the current column)
  %\includegraphics[width=.95\linewidth, bb=39 696 126 756]{figures/definition3.eps}
   \includegraphics[width=\linewidth]{ShapeNet1-eps-converted-to.pdf}
  % replacing the above command with the one below will explicitly set
  % the bounding box of the PS figure to the rectangle (xl,yl),(xh,yh).
  % It will also prevent LaTeX from reading the PS file to determine
  % the bounding box (i.e., it will speed up the compilation process)
  % \includegraphics[width=.95\linewidth, bb=39 696 126 756]{sampleFig}
  %
  %
  %%\vspace{-0.35in}
\caption{\label{fig:ShapeNet11}Visual comparison with COcc~\cite{DBLP:conf/eccv/PengNMP020} and IMLS~\cite{Liu2021MLS} in surface reconstruction under ShapeNet.}
%%\vspace{-0.28in}
\end{figure}

\begin{figure}[htb]
  \centering
  % the following command controls the width of the embedded PS file
  % (relative to the width of the current column)
  %\includegraphics[width=.95\linewidth, bb=39 696 126 756]{figures/definition3.eps}
   \includegraphics[width=\linewidth]{ShapeNet_Supp2-eps-converted-to.pdf}
  % replacing the above command with the one below will explicitly set
  % the bounding box of the PS figure to the rectangle (xl,yl),(xh,yh).
  % It will also prevent LaTeX from reading the PS file to determine
  % the bounding box (i.e., it will speed up the compilation process)
  % \includegraphics[width=.95\linewidth, bb=39 696 126 756]{sampleFig}
  %
  %
  %%\vspace{-0.35in}
\caption{\label{fig:ShapeNet2}Visual comparison with COcc~\cite{DBLP:conf/eccv/PengNMP020} and IMLS~\cite{Liu2021MLS} in surface reconstruction under ShapeNet.}
%%\vspace{-0.28in}
\end{figure}

We also highlight our performance on point denoising and surface reconstructions on a large scale real scan in our video.

\section{Point Cloud Denoising}
\begin{figure*}[htb]
  \centering
  % the following command controls the width of the embedded PS file
  % (relative to the width of the current column)
  %\includegraphics[width=.95\linewidth, bb=39 696 126 756]{figures/definition3.eps}
   \includegraphics[width=\linewidth]{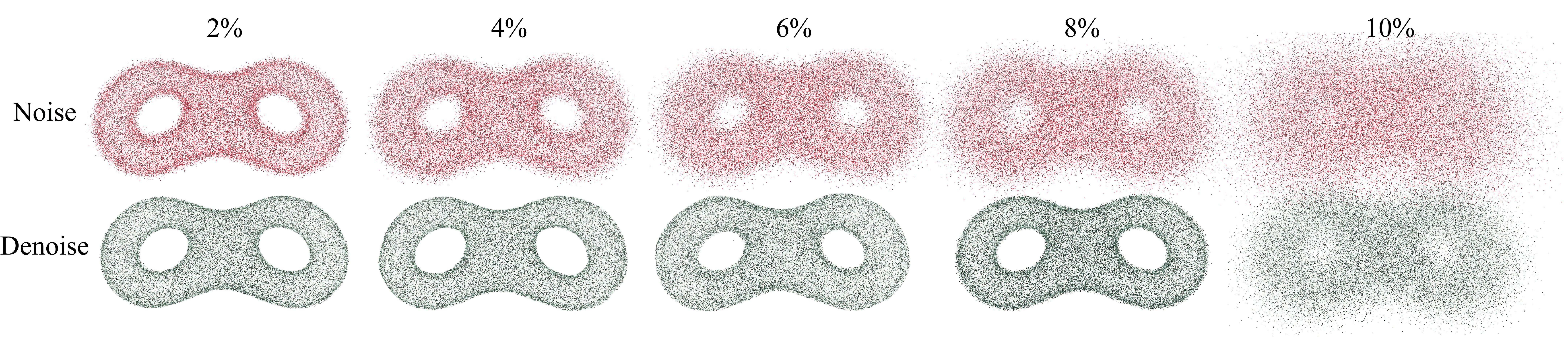}
  % replacing the above command with the one below will explicitly set
  % the bounding box of the PS figure to the rectangle (xl,yl),(xh,yh).
  % It will also prevent LaTeX from reading the PS file to determine
  % the bounding box (i.e., it will speed up the compilation process)
  % \includegraphics[width=.95\linewidth, bb=39 696 126 756]{sampleFig}
  %
  %
  %%\vspace{-0.35in}
\caption{\label{fig:noises}Point clouds denoising with large noises.}
%%\vspace{-0.28in}
\end{figure*}

Additionally, we visualize our results with larger noises which we use to learn an SDF in point cloud denoising in Fig.~\ref{fig:noises}. We tried noises with different variances including $\{2\%,4\%,6\%,8\%,10\%\}$. We can see that our method can reveal accurate geometry with large noises. While our method may fail if the noises are too large to observe the structures, such as the variance of 10 percent. Note that variances larger than 3 percent are not widely used in evaluations in previous studies.

\section{Results on KITTI}

\begin{figure}[htb]
  \centering
  % the following command controls the width of the embedded PS file
  % (relative to the width of the current column)
  %\includegraphics[width=.95\linewidth, bb=39 696 126 756]{figures/definition3.eps}
   \includegraphics[width=\linewidth]{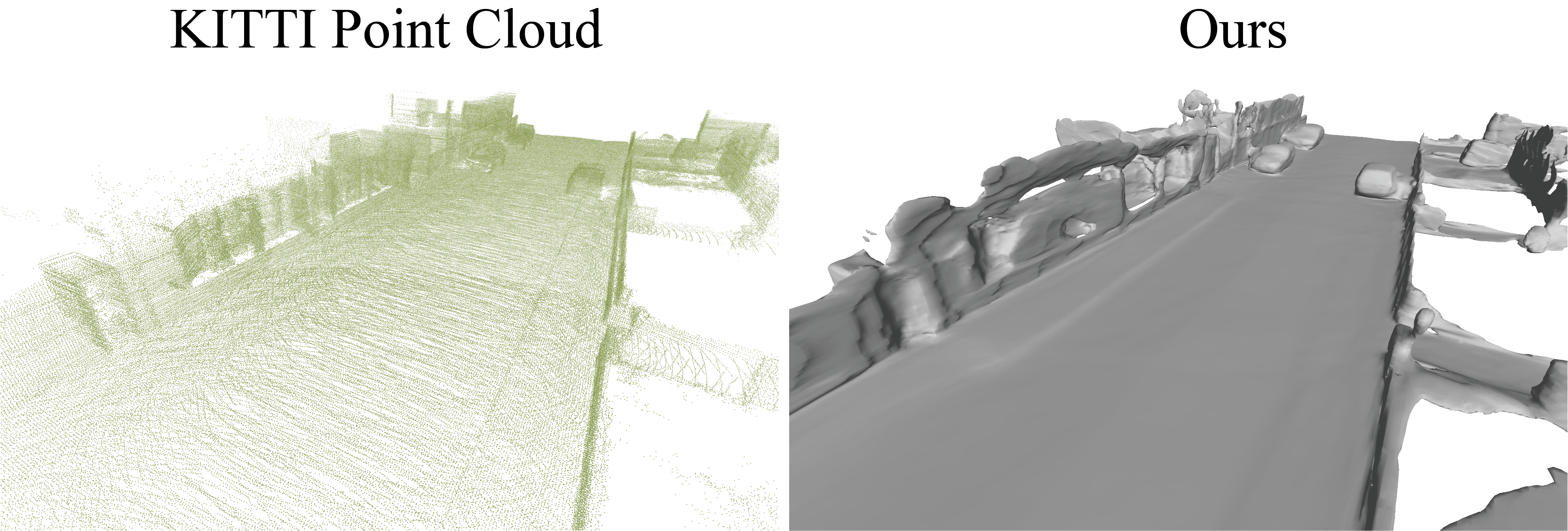}
  % replacing the above command with the one below will explicitly set
  % the bounding box of the PS figure to the rectangle (xl,yl),(xh,yh).
  % It will also prevent LaTeX from reading the PS file to determine
  % the bounding box (i.e., it will speed up the compilation process)
  % \includegraphics[width=.95\linewidth, bb=39 696 126 756]{sampleFig}
  %
  %
  %%\vspace{-0.35in}
\caption{\label{fig:KITTI}Reconstruction on a real scan from KITTI.}
%%\vspace{-0.28in}
\end{figure}

Additionally, we report our reconstruction on a road from KITTI in Fig.~\ref{fig:KITTI}. Our method can also reconstruct plausible and smooth surfaces from a single real scan containing sparse and noisy points, please see our reconstruction

\section{Computational Complexity}
We report our computational complexity in the following table. We report numerical comparisons with the latest overfitting based methods including NeuralPull (NP) and PCP using different point numbers including $\{20K,40K,80K,160K\}$ in Tab.~\ref{table:time}, where all methods search the nearest neighbors for queries online. NerualPull does not use learned priors while PCP uses learned priors parameterized by a neural network, both of which require the nearest neighbor search as ours. We report the time used to train these methods in 50K iterations. The comparisons indicate that our method uses less storage and less time than its counterparts.

%|       Time/GPU Memory     |      20K       |      40K       |      80K       |     160K       |
%|:-------------------------:|:--------------:|:--------------:|:--------------:|:--------------:|
%| NP                       |  12min/**1.5G**  |  15min/2.3G  |  19min/4.1G  |  33min/**8.0G**   |
%| PCP                       |  14min/1.9G     |  18min/2.7G     |  22min/4.6G     |  35min/8.4G     |
%| Ours                     |  **10min**/**1.5G**  |  **12min**/**2.2G**  |  **15min**/**4.0G**  |  **21min**/**8.0G**   |

\begin{table}[!ht]
    \centering
    \caption{\label{table:time}Comparison of Computational Complexity.}
    \resizebox{\linewidth}{!}{
    \begin{tabular}{l|l|l|l|l}
    \hline
        Time/GPU Memory & 20K & 40K & 80K & 160K \\ \hline
        NP & 12min/\textbf{1.5}G & 15min/2.3G & 19min/4.1G & 33min/\textbf{8.0}G \\ \hline
        PCP & 14min/1.9G & 18min/2.7G & 22min/4.6G & 35min/8.4G \\ \hline
        Ours & \textbf{10}min/\textbf{1.5}G & \textbf{12}min/\textbf{2.2}G & \textbf{15}min/\textbf{4.0}G & \textbf{21}min/\textbf{8.0}G \\ \hline
    \end{tabular}}
\end{table}

Since NP and PCP can not handle noises well, their reconstructions contain severe artifacts on the surface. While our method can handle that well. Please see more numerical comparisons with these methods in our paper. In addition, our results may get more improvements if we train our method more iterations.

\section{Ablation Studies}
\noindent\textbf{Number of Noisy Point Clouds. }We report additional ablation studies to explore the effect of the number of noisy point clouds in all the three tasks including point cloud denoising, point cloud upsampling, and surface reconstruction under the PU test set below. We can see we achieve the best performance with 200 noisy point clouds in all tasks, and the improvement over 100 point clouds is small. So we used 200 to report our results with multiple noisy point clouds in our paper.

%\begin{table*}[tb]
%\centering
%    \caption{Number of training iterations under PU.}
%    \begin{tabular}{c|c|c|c|c|c}
%     \hline
%          Iterations $\times 10^4$& Metric & 40 & 60 & 80 & 100 \\
%     \hline
%       \multirow{2}{*}{Denoise} & L2CD$\times10^4$&4.887&4.364&\textbf{4.221}&4.224\\
%       &P2M$\times10^4$&2.032&1.885&\textbf{1.847}&1.849\\
%       \hline
%       \multirow{2}{*}{Reconstruction} & L2CD$\times10^4$&4.971&4.624&\textbf{4.355}&4.330\\
%       &P2M$\times10^4$&2.312&1.985&\textbf{1.877}&1.991\\
%       \hline
%       \multirow{2}{*}{UpSampling} & L2CD$\times10^4$&4.968&4.447&\textbf{4.272}&4.342\\
%       &P2M$\times10^4$&2.211&1.954&\textbf{1.897}&1.991\\
%       \hline
%   \end{tabular}
%   \label{table:NOX15}
%\end{table*}

\noindent\textbf{Point Density. }We report the effect of point density in all the three tasks including point cloud denoising, point cloud upsampling, and surface reconstruction under the PU test set below. We learn an SDF from a single noisy point cloud. With more noises, our method can achieve better performance in all the three tasks.

\begin{table*}[htb]
\centering
\caption{Effect of Density $D$ of point Cloud under PU.}
%\resizebox{\linewidth}{!}{
    \begin{tabular}{c|c|c|c|c|c|c|c|c}
     \hline
          $D$& Metric & 1K & 2K & 5K & 10K & 20K & 50K &100K \\
     \hline
       \multirow{2}{*}{Denoise} & L2CD$\times10^4$&5.168&5.098&4.850&4.221&2.312&1.654 &\textbf{1.543}\\
       &P2M$\times10^4$&2.223&2.179&2.097&1.847&1.229&0.972 &\textbf{0.959}\\
       \hline
       \multirow{2}{*}{Reconstruction} & L2CD$\times10^4$&5.445&5.283&4.981&4.355&2.388&1.691 &\textbf{1.579}\\
       &P2M$\times10^4$&2.330&2.212&2.159&1.877&1.292&0.998 &\textbf{0.982}\\
       \hline
       \multirow{2}{*}{UpSampling} & L2CD$\times10^4$&5.281&5.187&4.984&4.272&2.392&1.682 &\textbf{1.561}\\
       &P2M$\times10^4$&2.398&2.212&2.167&1.897&1.289&0.997 &\textbf{0.973}\\
       \hline
   \end{tabular}%}
   \label{table:NOX17}
\end{table*}

\noindent\textbf{One Observation vs. Multiple Observations. }Since our method can learn from multiple observations and single observation, we investigate the effect of learning from these two training settings. Here, we combine multiple noisy observations into one noisy observation by concatenation, where we keep the total number of points the same. Table.~\ref{table:NOX18} indicates that there is almost no performance difference with these two training settings. The reason i

\begin{table}[htb]
\centering
\caption{Effect of mixing multiple noise point clouds under PU.}
%\resizebox{\linewidth}{!}{
    \begin{tabular}{c|c|c|c}
     \hline
          Strategy& Metric & Mixing & W/O Mixing  \\
     \hline
       \multirow{2}{*}{Denoise} & L2CD$\times10^4$&4.244&\textbf{4.221}\\
       &P2M$\times10^4$&1.851&\textbf{1.847}\\
       \hline
       \multirow{2}{*}{Reconstruction} & L2CD$\times10^4$&\textbf{4.315}&4.355\\
       &P2M$\times10^4$&\textbf{1.831}&1.877\\
       \hline
       \multirow{2}{*}{UpSampling} & L2CD$\times10^4$&4.299&\textbf{4.272}\\
       &P2M$\times10^4$&\textbf{1.897}&\textbf{1.897}\\
       \hline
   \end{tabular}%}
   \label{table:NOX18}
\end{table}

\section{Optimization Visualization}
We visualize the optimization process in our video. We visualize the noisy points matched by EMD for each query in each epoch. In addition, we also visualize the denoised points using the gradient in the learned SDF in different epochs.

%\section{Code}
%We release a demonstration code as a part of supplemental materials.

\section{Proof}

We proof Theorem 1 in our submission in the following.

\noindent\textbf{Theorem 1. }\textit{Assume there was a clean point cloud $\bm{G}$ which is corrupted into observations $S=\{\bm{N}_i\}$ by sampling a noise around each point of $\bm{G}$. If we leverage EMD as the distance metric $L$ defined in Eq.~(\ref{eq:4}), and learn a point cloud $\bm{G}'$ by minimizing the EMD between $\bm{G}'$ and each observation in $S$, i.e., $\min_{\bm{G}'}\sum_{\bm{N}_i\in S}L(\bm{G}',\bm{N}_i)$, then $\bm{G}'$ converges to the clean point cloud $\bm{G}$, i.e., $L(\bm{G},\bm{G}')=0$.}
\vspace{-0.1in}

\begin{equation}
\label{eq:4}
\begin{aligned}
L(\bm{G},\bm{G}')=\min_{\phi:\bm{G}\to\bm{G}'}\sum_{\bm{g}\in\bm{G}}||\bm{g}-\phi(\bm{g})\|_2,
\end{aligned}
\end{equation}
\vspace{-0.15in}

\noindent where $\phi$ is a one-to-one mapping.

\noindent\textbf{Proof: }Suppose each corrupted observation $\bm{N}_i$ in the set $S=\{\bm{N}_i|i\in[1,N]\}$ is formed by $m$ points, and $\bm{N}_i=\{n_i^k|k\in[1,m],m\ge 1\}$. With the same assumption, either $\bm{G}$ or $\bm{G}'$ is also formed by $m$ points, $\bm{G}=\{g^k|k\in[1,m],m\ge 1\}$, $\bm{G}'=\{g'^k|k\in[1,m],m\ge 1\}$. Assuming each noise $n_i^k$ is corrupted from the clean $g^k$, we leverage this assumption to justify the correctness of our proof. $L(\bm{G}',S)=\sum_{\bm{N}_i\in S}L(\bm{G}',\bm{N}_i)$.

$(a)$ When $m=1$, this is similar to Noise2Noise~\cite{noise2noise},

\begin{equation}
\begin{split}
\label{eq:Exp}
L(\bm{G}',S)=&\sum_{i=1}^{N}(g'^1-n_i^1)^2. \\
\frac{\partial L(\bm{G'},S)}{\partial \bm{G'}}&=2\sum_{i=1}^{N}(g'^1-n_i^1). \\
\frac{\partial L(\bm{G'},S)}{\partial \bm{G'}}=0 \rightarrow &g'^1 = 1/N\sum_{i=1}^{N}n_i^1.
\end{split}
\end{equation}

Since $S=\{\bm{N}_i\}$ is a set corrupted from the clean point cloud $\bm{G}$, $g^1=1/N\sum_{i=1}^{N}n_i^1$. Furthermore, we also get $g'^1=g^1$.

From Eq.~(\ref{eq:Exp}), we can also get the following conclusion,
\begin{equation}
\label{eq:min}
\mathop{min}\limits_{\bm{G'}}L(\bm{G'},S)\leftrightarrow \bm{G'}=\mathbb{E}(\bm{\phi}(\bm{G'})),
\end{equation}

\noindent where $\bm{\phi}=\{\phi_i|i\in[1,N]\}$ is a set of one-to-one mapping $\phi_i$ which maps $\bm{G'}$ to each corrupted observation $\bm{N}_i$ in $S$.

$(b)$ When $m\ge2$, assuming that we know which noisy point $n_i^k$ on each point cloud $\bm{N}_i$ is corrupted from the clean point $g^k$. We regard the correspondence $c_i$ between $\{n_i^k|i\in[1,N]\}$ and $g^k$ as the ground truth, so that we can verify the correctness of our following proof. Note that we did not use this assumption in the proof process. So, we can represent the correspondence using the following equation,

\begin{equation}
\mathbb{E}(n(k))=1/N\sum_{i=1}^{N}n_i^k=g^k,
\end{equation}

\noindent where $n(k)=\{n_i^k|i\in[1,N]\}$.

As defined before, $\phi_i$ is the one-to-one mapping established in the calculation of EMD between $\bm{G}'$ and $\bm{N}_i$. Therefore, the distance between $\bm{G}'$ and noisy point cloud set $S$ is, $L(\bm{G}',S)=\sum_{k=1}^{m}(\sum_{i=1}^{N}((g'^k-\phi_i(g'^k))^2))$,

There are two cases. One is that the one-to-one mapping $\phi_i$ is exactly the correspondence ground truth $c_i$. The other is that $\phi_i$ is not the correspondence ground truth.

Case $(1)$: When $\phi_i(g'^k)=n_i^k$, $i\in[1,N]$, this is consistent with $(a)$, so the Theorem 1 gets proved.

Case $(2)$: When $\phi_i(g'^k)\neq n_i^k$, assuming $\phi_i(g'^k)=n_i^{a_{k,i}}$, $A_k=\{n_i^{a_{k,i}}|i\in[1,N]\}$, $A_k$ is a set corresponding to $g'^k$. When minimizing $L(\bm{G}',S)=\sum_{k=1}^{m}\sum_{i=1}^{N}(g'^k-\phi_i(g'^k))^2$, according to Eq.~(\ref{eq:min}), $g'^k=\mathbb{E}(\phi_i(g'^k))$, so $Var(A_k)=1/N\sum_{i=1}^{N}((g'^k-\mathbb{E}(\phi_i(g'^k)))^2)$.
When $m=2$, $\mathop{min}\limits_{\bm{G'}}L(\bm{G'},S)=\min(Var(A_1)+Var(A_2))$. We assume $A_1=n_s^1+n_{cs}^2$ to simply the following proof, where $s$ is a subset of set $[1,N]$, ${cs}$ is the complement of set $s$, so $A_2=n_s^2+n_{cs}^1$. Assuming $\mathbb{E}(A_1)=g^1+\Delta$, $\Delta$ is the point offset of $g^1$, because of $\mathbb{E}(A_1)+E(A_2)=g^1+g^2$, so $\mathbb{E}(A_2)=g^2-\Delta$,
%\clearpage

\begin{equation}
\begin{split}
L(\bm{G'},S)\\
=&(Var(A_1)+Var(A_2))\\
   =& \mathbb{E}(A_1-(g^1+\Delta))^2+\mathbb{E}(A_2-(g^2-\Delta))^2 \\
    =& 1/N(\sum_{i=1}^{N}(n_i^{a_{1,i}})^2+\sum_{i=1}^{N}(n_i^{a_{2,i}})^2+N(g^1+\Delta)^2\\
    &+N(g^2-\Delta)^2-2\sum_{i=1}^{N}n_i^{a_{1,i}}(g^1+\Delta)\\
    &-2\sum_{i=1}^{N}n_i^{a_{2,i}}(g^2-\Delta)) \\
    =& \mathbb{E}((n(1))^2)+\mathbb{E}((n(2))^2)+\mathbb{E}^2(n(1))+\\
    &\mathbb{E}^2(n(2))+2\Delta^2+2g^1\Delta-2g^2\Delta-\\
    &2/N(g^1\sum_{i=1}^{N}n_i^{a_{1,i}}+g^2\sum_{i=1}^{N}n_i^{a_{2,i}}+\\
    &\Delta\sum_{i=1}^{N}n_i^{a_{1,i}}-\Delta\sum_{i=1}^{N}n_i^{a_{2,i}}) \\
    =&\mathbb{E}((n(1))^2)+\mathbb{E}((n(2))^2)+\mathbb{E}^2(n(1))+\\
    &\mathbb{E}^2(n(2))+2/N(\Delta(n_s^1+n_{cs}^1)-\Delta(n_s^2+n_{cs}^2)\\
    &-\Delta(n_s1+n_{cs}^2)+\Delta(n_{cs}^1+n_s^2)- \\
    &g^1\sum_{i=1}^{N}n_i^{a_{1,i}}-g^2\sum_{i=1}^{N}n_i^{a_{2,i}})\\
    =& \mathbb{E}((n(1))^2)+\mathbb{E}((n(2))^2)+\mathbb{E}^2(n(1))+\\
    &\mathbb{E}^2(n(2))+2\Delta^2+{2\Delta}/N(2n_{cs}^1-\\
    &2n_{cs}^2)-2/N(g^1N(g^1+\Delta)+g^2N(g^2-\Delta)) \\
    =& \mathbb{E}((n(1))^2)+\mathbb{E}((n(2))^2)-\mathbb{E}^2(n(1))-\\
    &\mathbb{E}^2(n(2))+2\Delta^2+\\
    &{2\Delta}/N(2n_{cs}^1-2n_{cs}^2-n_{cs}^1-n_s^1+n_{cs}^2+n_s^2) \\
    =& \mathbb{E}((n(1))^2)+\mathbb{E}((n(2))^2)-\mathbb{E}^2(n(1))-\\
    &E^2(n(2))+2\Delta(g^2-g^1)-2\Delta^2 \\
    =& Var(n(1))+Var(n(2))+2\Delta(g^2-g^1)-2\Delta^2 \\
\end{split}
\end{equation}

Because the first two terms of the formula are constants, the entire formula becomes a quadratic formula, so when $\Delta=0$ or $\Delta=g^2-g^1$, the value of $L(\bm{G'},S)$ is minimized. $\Delta=0$ is consistent with Case $(1)$. $\Delta=g^2-g^1$, $\phi_i(g^{1})=n_i^2$, $\phi_i(g^{2})=n_i^1$, this is also the same correspondence as the ground truth, so Theorem 1 gets proved. When $m\textgreater2$. We can extend the proof from the two sets $A_1$ and $A_2$ to multiple sets $A_1,A_2,\cdots,A_m$, and the proof process is similar to the above.

\begin{table*}[htb]
\centering
\caption{Effect of batch size $B$ under PU.}
    \begin{tabular}{c|c|c|c|c|c|c|c|c}
     \hline
          $B$ & Metric & 1 & 2 & 10 & 20 & 50 & 100&200\\
     \hline
       \multirow{2}{*}{Denoise}&L2CD$\times10^4$&4.976&4.898&4.665&4.558&4.432	&4.224&\textbf{4.221}\\
       &P2M$\times10^4$&2.132&2.079&1.997&1.996&1.899&\textbf{1.847}&\textbf{1.847}\\
      \hline
       \multirow{2}{*}{Reconstruction}&L2CD$\times10^4$&5.102&4.995&4.795&4.599&4.456&4.369 &\textbf{4.355}\\
       &P2M$\times10^4$&2.423&2.217&2.007&2.001&1.978&1.886&\textbf{1.877}\\
      \hline
       \multirow{2}{*}{UpSampling}&L2CD$\times10^4$&4.988&4.886&4.687&4.574&4.461	&4.328&\textbf{4.272}\\
       &P2M$\times10^4$&2.152&2.082&2.001&1.997&1.977&1.919
 &\textbf{1.897}\\
      \hline
   \end{tabular}
   \label{table:NOX14}
\end{table*}

\begin{table*}[htb]
\centering
%\caption{F-Score comparison under ShapeNet.}  % ????????
\resizebox{\linewidth}{!}{
%\makebox[\linewidth]{
    \begin{tabular}{c|cccccccccc}  % ?????
     \toprule
     %\multirow{2}{*}{category}&\multicolumn{9}{c}{F-Score} \\
     %&PSR&PSGN&R2N2&AtlasNet&ConvONet&SAP&OCNNC&IMLS&\textbf{Ours}\\
     &PSR&PSG&R2N2&Atlas&COcc&SAP&OCNN&IMLS&POCO&\textbf{Ours}\\
     \hline
     airplane&0.551&0.476&0.382&0.827&0.965&0.981&0.810&0.992&0.994&\textbf{0.995}\\
     bench&0.430&0.266&0.431&0.786&0.965&0.979&0.800&0.986&0.988&\textbf{0.993}\\
     cabinet&0.728&0.137&0.412&0.603&0.955&0.975&0.789&0.981&0.979&\textbf{0.996}\\
     car&0.729&0.211&0.348&0.642&0.849&0.928&0.747&0.952&0.946&\textbf{0.964}\\
     chair&0.473&0.152&0.393&0.629&0.939&0.979&0.799&0.982&0.985&\textbf{0.993}\\
     display&0.544&0.175&0.401&0.727&0.971&0.990&0.811&0.994&0.994&\textbf{0.998}\\
     lamp&0.586&0.204&0.333&0.562&0.892&0.959&0.800&0.979&0.975&\textbf{0.990}\\
     speaker&0.731&0.107&0.405&0.516&0.892&0.957&0.779&0.963&0.964&\textbf{0.977}\\
     rifle&0.590&0.615&0.381&0.877&0.980&0.990&0.826&0.996&0.998&\textbf{0.998}\\
     sofa&0.712&0.184&0.427&0.717&0.953&0.982&0.801&0.987&0.989&\textbf{0.992}\\
     table&0.442&0.158&0.404&0.692&0.967&0.986&0.801&0.987&0.991&\textbf{0.992}\\
     telephone&0.674&0.317&0.484&0.867&0.989&0.997&0.825&0.998&0.998&\textbf{0.999}\\
     vessele&0.771&0.363&0.394&0.7757&0.931&0.974&0.809&0.987&0.989&\textbf{0.997}\\
     \hline
     mean&0.612&0.259&0.400&0.708&0.942&0.975&0.800&0.983&0.984&\textbf{0.991}\\
     \toprule
\end{tabular}%}
   }%\vspace{-0.15in}
   \caption{F-Score comparison under ShapeNet.}
   \label{table:shapenet2app}%\vspace{-0.2in}
\end{table*}
%%%%%%%%%%%%%%%%%%%%%%%%%%%%%%%%%%%%%%%%%%%%%%%%%%%%%%%%%%%%%%%%%%%%%%%%%%%%%%%%
%%%%%%%%%%%%%%%%%%%%%%%%%%%%%%%%%%%%%%%%%%%%%%%%%%%%%%%%%%%%%%%%%%%%%%%%%%%%%%%%

\end{document}